\documentclass{SCIS2023}
\usepackage{graphicx}
\usepackage{color}

\begin{document}
\ArticleType{RESEARCH PAPER}
\Year{2022}
\Month{}
\Vol{}
\No{}
\DOI{}
\ArtNo{}
\ReceiveDate{}
\ReviseDate{}
\AcceptDate{}
\OnlineDate{}

\title{Reducing Idleness in Financial Cloud Services via Multi-objective Evolutionary Reinforcement Learning based Load Balancer}{}

\author[1,2]{Peng YANG}{yangp@sustech.edu.cn, Peng YANG}
\author[2,3]{Laoming ZHANG}{}
\author[4]{Haifeng LIU}{}
\author[2,5]{Guiying LI}{}

\AuthorMark{Author A}

\AuthorCitation{Author A, Author B, Author C, et al}


\address[1]{Department of Statistics and Data Science, Southern University of Science and Technology, Shenzhen {\rm 518055}, China}
\address[2]{Guangdong Provincial Key Laboratory of Brain-Inspired Intelligent Computation, \\Department of Computer Science and Engineering, Southern University of Science and Technology, Shenzhen {\rm 518055}, China}
\address[3]{Academy for Advanced Interdisciplinary Studies, Southern University of Science and Technology, Shenzhen {\rm 518055}, China}
\address[4]{Guangdong OPPO Mobile Telecommunications Corp., Ltd, Shenzhen {\rm 518052}, China}
\address[5]{Research Institute of Trustworthy Autonomous Systems, Southern University of Science and Technology, Shenzhen {\rm 518055}, China}

\abstract{In recent years, various companies have started to shift their data services from traditional data centers to the cloud. One of the major motivations is to save on operational costs with the aid of cloud elasticity. This paper discusses an emerging need from financial services to reduce the incidence of idle servers retaining very few user connections, without disconnecting them from the server side. This paper considers this need as a bi-objective online load balancing problem. A neural network based scalable policy is designed to route user requests to varied numbers of servers for the required elasticity. An evolutionary multi-objective training framework is proposed to optimize the weights of the policy. Not only is the new objective of idleness is reduced by over 130\% more than traditional industrial solutions, but the original load balancing objective itself is also slightly improved. Extensive simulations with both synthetic and real-world data help reveal the detailed applicability of the proposed method to the emergent problem of reducing idleness in financial services.}


\maketitle

\section{Introduction}
Elasticity forms a cornerstone of cloud computing that has helped it succeed in the past decade\cite{Elasticity}. In many fields, the number of concurrent users of a cloud service system may vary significantly over time, while the cloud system can reduce the idle servers elastically when the users become fewer, and vice versa \cite{Qu2019Auto}. The concept of idle servers refers to those servers with few user tasks, in which most of the hardware resources, e.g., the CPU, are running to no purpose. Despite the environmental issues, according to \cite{qureshi2009cutting}, the energy consumption cost may amount to 30\%–50\% of the operational cost of those large-scale data centers built by companies such as Google, Microsoft, and Facebook, which is a huge waste of money. Those costs will eventually be transferred to the tenants who pay for and use the cloud servers \cite{shastri2016transient}. In financial scenarios (especially in stock trading), it is common for the number of concurrent users at the close of the market to be over 10 times smaller than that at the opening \cite{Stoll2006Elec}. Traditional financial services focus more on stability than elasticity, and now the cloud native offers a new possibility for achieving both in the same architecture \cite{li2019cloud}.

Unfortunately, no industrial solution has been available to readily reduce idle servers in the financial cloud services as it is still no trivial task. Ideally, if a server serves zero user connections, it can be shut down to reduce idleness. In practice, when the user concurrency decreases markedly, the servers that become idle may still retain a small number of user connections. Other services such as the entertainment and office sectors can simply disconnect all these users and re-connect them to the busy servers quickly. In this regard, these idle servers become effectively empty and thus, can be shut down to save costs. However, in financial services, the availability of the connections is extremely sensitive, as even a millisecond of disconnection may cause a heavy loss of investment and damage to the financial market \cite{easley2011microstructure}. This means that an idle server in the financial services sector must wait passively for all its users (even though very few) to disconnect and then can be shut down afterward, while the waiting time is a direct waste of the idle servers.

We propose that the above issue is mainly caused by the improper routing decisions of the user requests over the servers. Specifically, given that a server can only be shut down when it serves no connection, we know that the shutdown time of a server is decided by the latest disconnection time among all its user connections. Once we can route the user connections with a similar disconnection time in the same server, this server can be shut down in time and the wasted idleness can be greatly reduced. Based on this, the present paper focuses on the load balancer module within the cloud computing architecture, as it is responsible for routing user requests to different servers to construct connections.

As its name suggests, the traditional routing algorithms inside the industrial load balancers mainly focus on minimizing the load imbalance, i.e., the standard deviation of the workloads among multiple servers. The performance of those servers with heavy workloads can be very poor and will impact the whole system negatively \cite{mirobi2019dynamic}. Furthermore, overloaded servers may fail much more easily. In the case of servers with light workloads, the hardware resources are thereby wasted \cite{kansal2012cloud}. Thus, the load balance has been a major problem in cloud computing \cite{Kumar2019issues,Chengzhi2017Alibaba}. By bearing the workloads, such as user connections, the hardware resources of the servers will be consumed and, thus, considered the indicators of the load balance. For modern computing systems, four basic types of resources are the CPU and memory (RAM) for computing, a disk for storage, and bandwidth (BW) for network communication. In this paper, we consider all of them to make the proposed method more general in terms of hardware resources. The representative routing algorithms are heuristics such as Round Robin, Random Routing, IP Hash method, the Least Connection, and their weighted variants \cite{shafiq2022load, mirobi2019dynamic, johora2022load, marinescu2017approach}. These algorithms are based on either some degree of randomness or some hardware indicators of the servers. None of these considers the behavior of users, i.e., the user connection duration. In this regard, while the users were routed to the servers for load balance, the remaining user connections at the close of the market will also be distributed randomly among the servers. This explains why, in real cases, the idle servers usually retain a few connections when the user concurrency decreases significantly. Therefore, although they have contributed greatly to enhancing the capacity (concurrent users) and the stability of the cloud services by balancing workloads, they may perform poorly on routing the user requests concerning further reducing idle servers without killing connections, as is required in financial services. Recently, intelligent routing algorithms have also been developed based on meta-heuristics and reinforcement learning \cite{sim2003ant, gures2022machine, brar2016meta, farag2021congestion}. Unfortunately, their targeted scenarios were quite different from the financial cloud and thus, did not encounter the idleness issue.

To address this issue, this paper aims to automatically learn the optimal policy for routing incoming user requests over the available servers so that the load imbalance and the idle time of all servers can be minimized simultaneously. For this purpose, when routing an incoming user request, we not only reorganize the traditional indicators of the servers along the time axis but also explicitly exploit the historical connection duration of that user as new features. Specifically, it is assumed that the connection duration of each user on different days follows a similar distribution. Thus, the connection duration of an incoming user can be predicted statistically. This assumption is reasonable because most of the users of financial services are from financial companies who usually have regular work schedules. Different degrees of the predictability of users are simulated in the experiments to support this assumption positively.
Due to the online decision-making nature of routing incoming requests, the whole routing process is modeled as a Reinforcement Learning (RL) problem with two objectives, namely, the load imbalance and the idle time of all available servers. Then, a parameter-sharing neural-network-based scalable routing policy is designed for two reasons: 1) to learn about the non-linear relationship between the statistically estimated connection duration and the states of the servers; 2) to deal with the elastic scenarios where the number of servers may change due to the autoscaling or servers’ failure. Considering that it is difficult to set proper weights for aggregating the two objectives, an evolutionary multi-objective algorithm-based training framework is proposed to optimize the parameters of the neural-network-based policy. Lastly, an action mask operator is designed to help train the policy stably. As a result, a Multi-objective Evolutionary Reinforcement Learning-based Load Balancer (MERL-LB) is proposed.

Extensive simulations show that a MERL-LB can outperform comparable algorithms significantly on the emerging task of reducing idleness without disconnecting user connections. The diverse Pareto optimal produced by the evolutionary multi-objective training framework offers various options for the users. Among the diverse policies, the idleness objective can be reduced by over 130\% in using the traditional methods while the load balancing objective itself is still slightly improved. Meanwhile, it has been shown that the evolutionary multi-objective training framework facilitates a much faster convergence rate over policy gradients \cite{schulman2017proximal}. The proposed scalable routing policy has also been verified successfully against a variety of servers and request loads, with the number of users ranging from 600 to 7500. Finally, an emerging sawtooth pattern is observed from the process of MERL-LB-based policies, which has been studied by comparing it to the patterns of traditional methods in detail to explain its rationality and may shed light on designing novel heuristics in the future. In a nutshell, MERL-LB is shown empirically to be a powerful solution to the emerging problem of reducing idleness and load imbalance in financial services.

The remainder of this work is as follows. Section 2 reviews the related works. Section 3 describes the problem scenarios. Section 4 first models this problem as an RL problem and then presents the scalable neural-network-based policy and its evolutionary multi-objective algorithm-based training framework. The simulations are conducted in Section 5 to verify the effectiveness of the proposed MERL-LB. The conclusions drawn from this work are discussed in the last section.

\section{Related works} \label{sec: related works}
The load balancer works between the users’ applications and the servers of the backend system. In cloud data services, the users’ requests come up in an online manner. The load balancer is expected to route each incoming user’s request to one of the available servers so that any single server in the data center will not be overloaded. And once the request is routed successfully, a connection is constructed between the user’s application and the routed server to bi-directionally transmit data.

Industrial routing algorithms mostly make their decisions with certain degrees of randomness or based on the states of the servers to keep load balance. Typical static routing algorithms include Random Routing, Round Robin, Weighted Round Robin, and IP Hash \cite{sajjan2017load}. Round Robin and Weighted Round Robin distribute requests according to a specific probabilistic distribution among servers \cite{marinescu2017approach}. IP Hash maps the IP address to the corresponding server using a hash algorithm. Static routing algorithms are generally simple to implement but lack the ability to adapt their probabilistic distributions to the changing characteristics of the network traffic. The other type of dynamic load balancing algorithms considers the servers’ states as input \cite{alakeel2010guide}. Examples are Least Connection \cite{shafiq2022load} and the Throttled method \cite{johora2022load}. Among them, the Least Connection dynamically assigns tasks based on the already served number of connections of each server. The Throttled method consists of an index table of available virtual machines and their states. Requests are allocated to the virtual machine that is available and has sufficient resources \cite{chen2017clb}. However, those methods are designed for the single objective of load balancing and cannot address the problem of reducing idleness without disconnecting user connections.

There are also some intelligent load balancers. Meta-heuristics based load balancers \cite{sim2003ant} are general derived from the behavior of natural evolution and iteratively search the optimal solution in the policy space. Unfortunately, those methods can hardly be applied to the underlying online real-time routing scenarios, as their iterative optimization process usually takes a relatively long time to find an optimal routing solution \cite{brar2016meta}. RL-based load balancers are also warmly studied for the purpose of real-time routing. The witnessed works mainly focus on the scenarios of Software Defined Networks \cite{gures2022machine} and Internet of Things networks \cite{farag2021congestion}. However, as the elasticity is not the major concern of those networks, those works do not encounter the same difficulty as in the financial scenarios of keeping connections alive while reducing idleness.

For cloud computing scenarios, the general load balance problems have been more intensively studied in the context of job shop scheduling \cite{afzal2019load, carrion2022kubernetes, Kashani2023Load} in the scheduler module in the backend. Honestly, the job scheduling problems and the connection routing problems generally follow the same vector bin-packing mathematical model \cite{trivella2016load, Kumar2019issues}, where a set of vector items are expected to be distributed among multiple bins so that their loads can be balanced. Unfortunately, we discuss as follows that those job scheduling methods cannot readily address the underlying problem of this work. The two major differences between them emerge from practice.

(1) A computing job is basically a sequence of predefined computing steps where any intermediate data can be easily saved as checkpoints with established techniques. In this regard, killing a job on one server and restarting it on another server can lose no data with extra techniques like Live Migration \cite{basu2019learn}. However, a user connection does not know what data will be delivered through it at a specific time, which means a disconnection will lose data that can hardly be recovered. In a nutshell, those job scheduling methods do not explicitly keep connections alive while reducing idleness.

(2) Though some online job scheduling works also consider objectives similar to reducing idleness, like minimizing slowdown \cite{zhu2022qos}, makespan \cite{zhang2023smaf} or job completion time \cite{wang2022solving}, they need the running duration of the jobs explicitly in advance. While in the connection routing problem, the online duration of any user connection is unknown when the request comes up.


To summarize, to our best knowledge, existing related works are not dedicated to solving the underlying bi-objective problem, i.e., load balancing and reducing idleness without actively disconnecting users. Thus, directly applying them may not be satisfactory.

\section{Problem description}
\begin{figure}[htbp]
\centering
\begin{minipage}[t]{0.44\linewidth}
\centering
\includegraphics[scale=0.47]{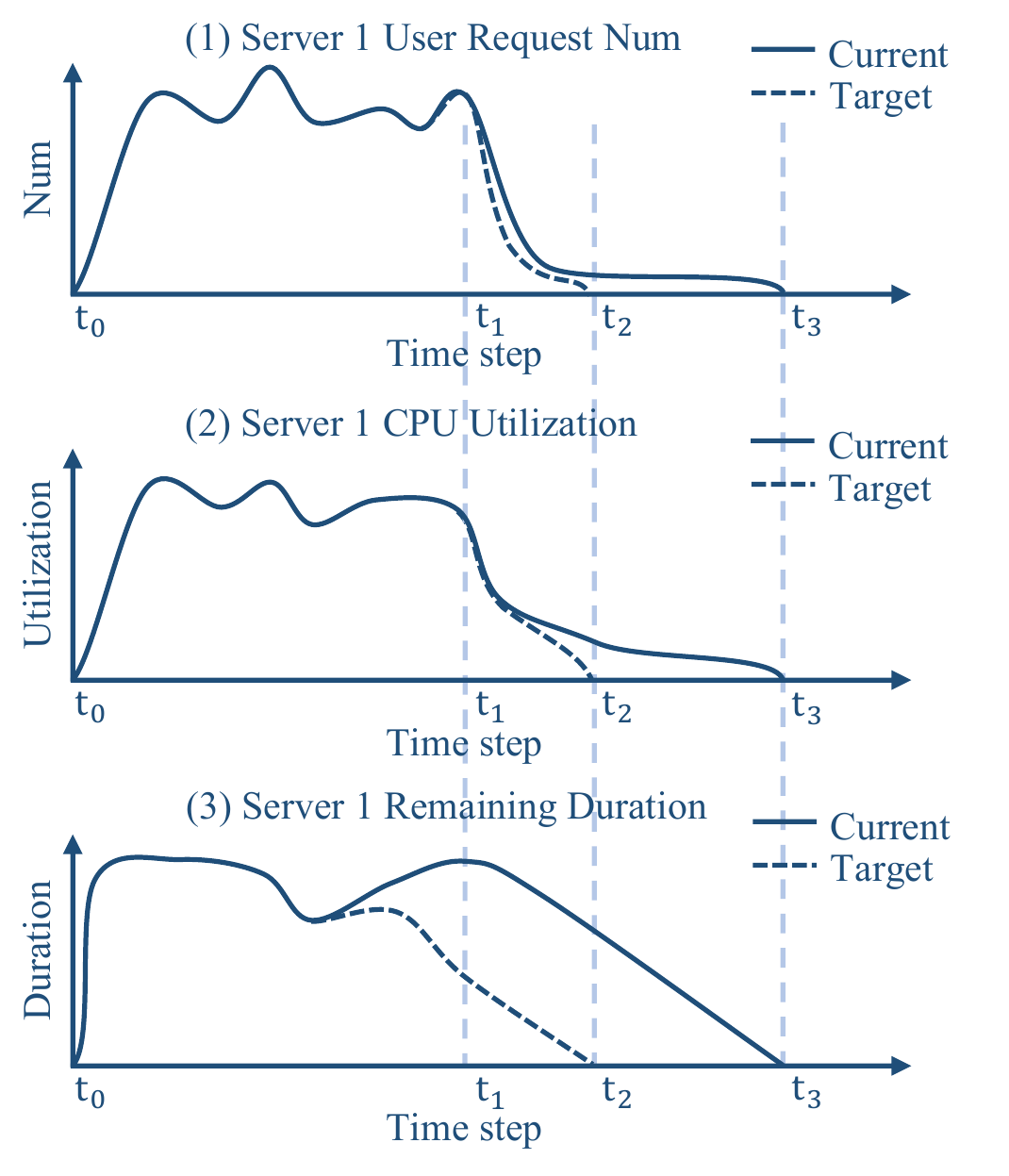}
\caption{A typical phenomenon of financial services that there exist very few connections from $t_2$ to $t_3$, which are targeted to be reduced in this work.}
\label{fig: a typical phenomenon of financial services}
\end{minipage}
\begin{minipage}[t]{0.55\linewidth}
\centering
\includegraphics[scale=0.47]{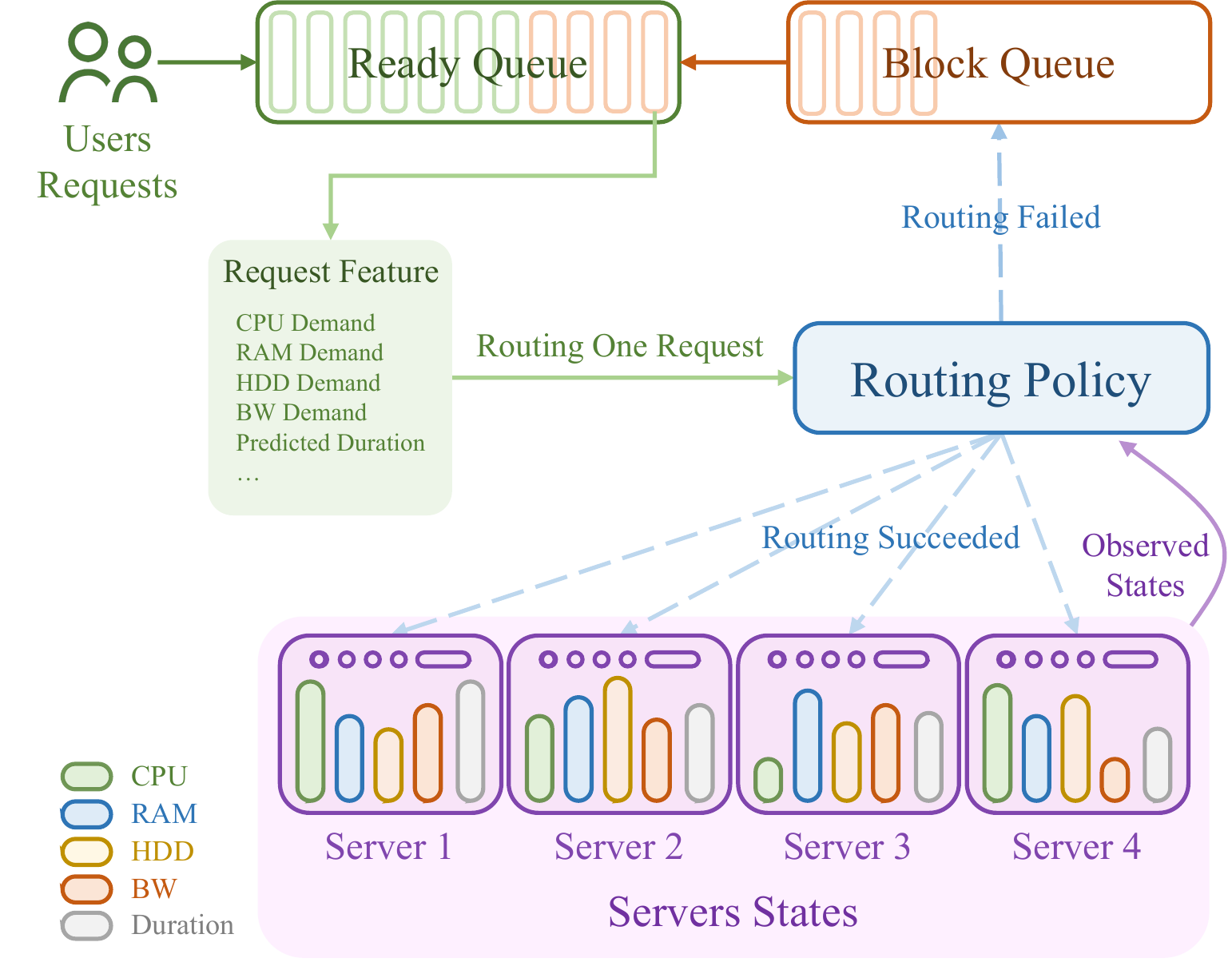}
\caption{The routing process of load balancer in financial services.}
\label{fig: routing process}
\end{minipage}
\end{figure}

In typical financial data services, e.g., the stock market data services, the user connections usually follow a quite regular mode. First, the number of user connections increases very rapidly at the market opening in the morning, and then decreases gradually when the o’clock is approaching to the market closing in the afternoon. Second, the duration of different user connections may vary significantly. Consequently, as shown in Fig.\ref{fig: a typical phenomenon of financial services}, the users usually disconnect from the data services at different times, leaving the servers running with a very low utilization rate but cannot be shutdown due to few alive connections on them, e.g., the duration from $t_2$ to $t_3$. This economic waste resulting from the server idleness can be significantly scaled up as the financial data services usually run on cloud computing infrastructures with large-scale servers. 


Fortunately, the duration of each user connection in different days often varies slightly as most users are staff from financial industries who often have regular schedules \cite{kandel2012enterprise}. This simply suggests that the disconnection time of the same user on different days can be predictive to some extent and maybe statistically modeled following a probabilistic distribution. With this special feature in the financial services, this work describes the problem of routing user connections to reduce idleness as follows. 

As shown in Fig.\ref{fig: routing process}, each incoming user request has five types of features, including the demand of CPU, memory (RAM), storage (HDD), bandwidth (BW) and the historical distribution of the connection duration of this user. The first four features are hardware requirements indicating how many resources the system needs to serve that connection and its related data services. The last feature emerges from the financial services and is, for the first time, considered in the load balancer. Each time a new request arrives, it first enters the ready queue and waits to be routed to one of the available servers by the routing policy. Once the user request is successfully routed to a server, a user connection between the user and the server can be constructed, and data services can be provided to the user through the connection. Otherwise, those requests that cannot be routed due to insufficient hardware resources will be tentatively placed in the block queue until the next round of routing. The routing policy works by optimally distributing the users’ requests to minimize the load imbalance of 4 types of resource utilization among servers while simultaneously reducing the idle time of the servers. 

In this paper, the above mentioned two objective functions are defined as follows. Suppose the load balancer continues serving for a period of timesteps. For the objective of load balancing, it is defined as the standard deviation of the utilization among 4 resource types:In this paper, the above mentioned two objective functions are defined as follows. Suppose the load balancer continues serving for a period of T timesteps. For the objective of load balancing, it is defined as the standard deviation of the utilization among 4 resource types:
\begin{equation}
    F_{\text {balance }}=\frac{1}{4 \cdot T} \sum_{t=0}^{T} \sum_{r=1}^{4} \sqrt{\frac{\sum_{i=1}^{N}\left(x_{r i}^{t}-\mu_{r}^{t}\right)^{2}}{N}},
    \label{equ: balance fitness}
\end{equation}
where $x_{ri}^t$ indicates the utilization of r-th resource type of the i-th server at the t-th timestep, and $\mu_r^t$ denotes the average of the r-th type resource utilization among all $N$ servers at the t-th timestep. For the objective of reducing idle time, instead of using the commonly-seen makespan of all the servers, we calculate the average remaining duration of the servers over all timesteps. The reason is that we want to achieve idle reduction at any time, not just at the last moment.
\begin{equation}
    F_{\text {idle }}=\frac{1}{T \cdot N} \sum_{t=0}^{T} \sum_{i=1}^{N} d_{i}^{t},
    \label{equ: idleness fitness}
\end{equation}
where $d_i^t$ is maximal remaining duration among all connections of the $i$-th server at the $t$-th timestep. Note that, the remaining duration of any connection cannot be known exactly before its disconnection. On the other hand, as discussed previously, according to the observation that each user’s connection duration on different official days changes slightly due to the regular schedule, we can predict the duration of each user at the current $t$-th timestep as the average of the user’s historical duration. This paper also empirically assesses the different impacts on the load balancer caused by different degree of predictability. Also note that, in the whole problem description, the system is not allowed to actively disconnect the user connections. Only passive disconnection from the user’s application is permitted.

\section{Method}
In this section, the RL-based problem formulation is first described, including the representation of actions, the RL states, and the rewards. Then, a neural network is designed to learn the non-linear relationship between the statistically estimated connection duration and the available servers. The architecture of this neural network is also designed for elastic scenarios with changing numbers of servers. After that, an evolutionary multi-objective algorithms based training framework is introduced for this scalable policy, in terms of both minimal load imbalance and server idleness. The action mask operator for helping stably train the policy is detailed lastly.

\subsection{Reinforcement learning based formulation}
In this RL model, each of the user’s requests in the ready queue is represented together with the utilization of all servers as the RL states. Based on the states, the policy outputs the action of routing the request to one of the servers to construct a user connection. 

\begin{figure}[htb]
  \centering
  \includegraphics[width=0.6\linewidth]{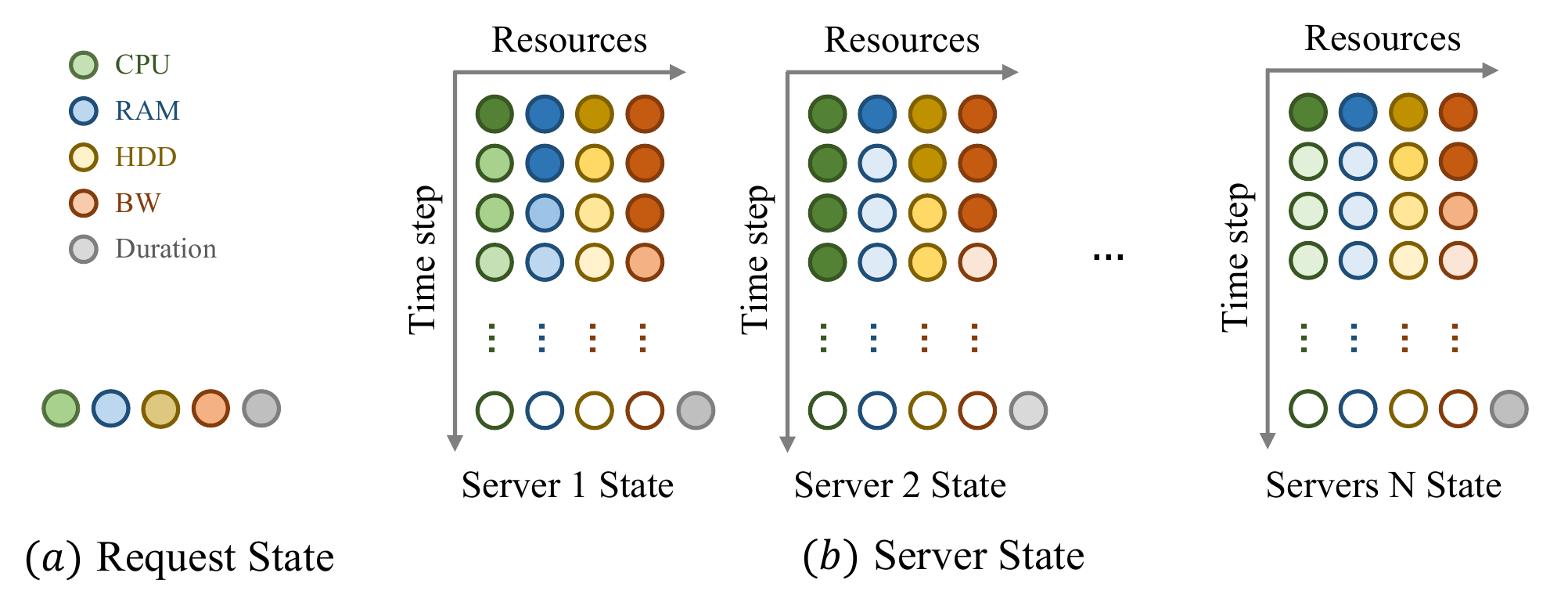}
  \caption{This work considers two types of states for RL: (a) the 5-dimensional states of each user request, (b) the 41N-dimensional server states include $N$ servers with 4 types of resources looking forward 10 timesteps into the future, and 1 maximal remaining duration.}
  \label{fig: state desgin}
\end{figure}

As shown in Fig.\ref{fig: state desgin}, the RL states are a conjunction of both the states of one user’s request and the states of $N$ servers. The state of each request is represented as a $5$-dimensional vector, including the demands of $4$ types of hardware resources and $1$ scalar value of the predicted duration, i.e., the average of the historical duration of that user. The state of each server is represented as a $(4h+1)$-dimensional vector, i.e., the utilization of $4$ types of resources along $h$ looking forward timesteps and $1$ scalar value of the maximal remaining duration among all connections in the future $h$-th timestep. Here the remaining duration of each connection is calculated as its predicted duration minus $h$ timesteps, and the utilization of each resource in some future timestep is calculated as the summation overof all alive connections at that moment. By using this feature along $h$ looking forward timesteps, the future states of the servers are involved to help the policy learn the optimal distribution of the requests for reducing idleness in the future. In the empirical studies, we simply set $h$ to $10$. As a result, the total length of the RL states vector is $5+41N$.

The RL model has $N+1$ actions, i.e., routing the request to one of the $N$ servers or the blocking queue if no proper server is available. For the reward function, as we employ the evolutionary multi-objective algorithm to train the policy, it is unnecessary to derive the intermediate rewards of each routing action. Instead, the two objectives in Eqs.\eqref{equ: balance fitness}-\eqref{equ: idleness fitness} are used as the accumulative rewards for each policy after the whole simulation process \cite{salimans2017evolution}.



\subsection{The scalable policy}\label{section: scalable policy}
To learn the optimal distribution of connections over servers, the neural network based policy is preferred due to its strong learning ability. The input of the neural network will be the above described $(5+41N)$-dimensional RL states, and the output of the neural network will be the $(N+1)$-dimensional RL actions. Therefore, the sizes of the input and output of the policy change with the number of available servers. Additionally, due to the elasticity of cloud computing, the number of available servers does usually change over time for a given data service. Besides, it is also important to generalize the trained policy to different data services that may be supported by different numbers of servers. For these purposes, the employed policy should be able to scale with the number of servers automatically. To this end, this paper proposes a parameter-sharing based neural network architecture for the underlying routing tasks.

\begin{figure}[htb]
  \centering
  \includegraphics[width=0.6\linewidth]{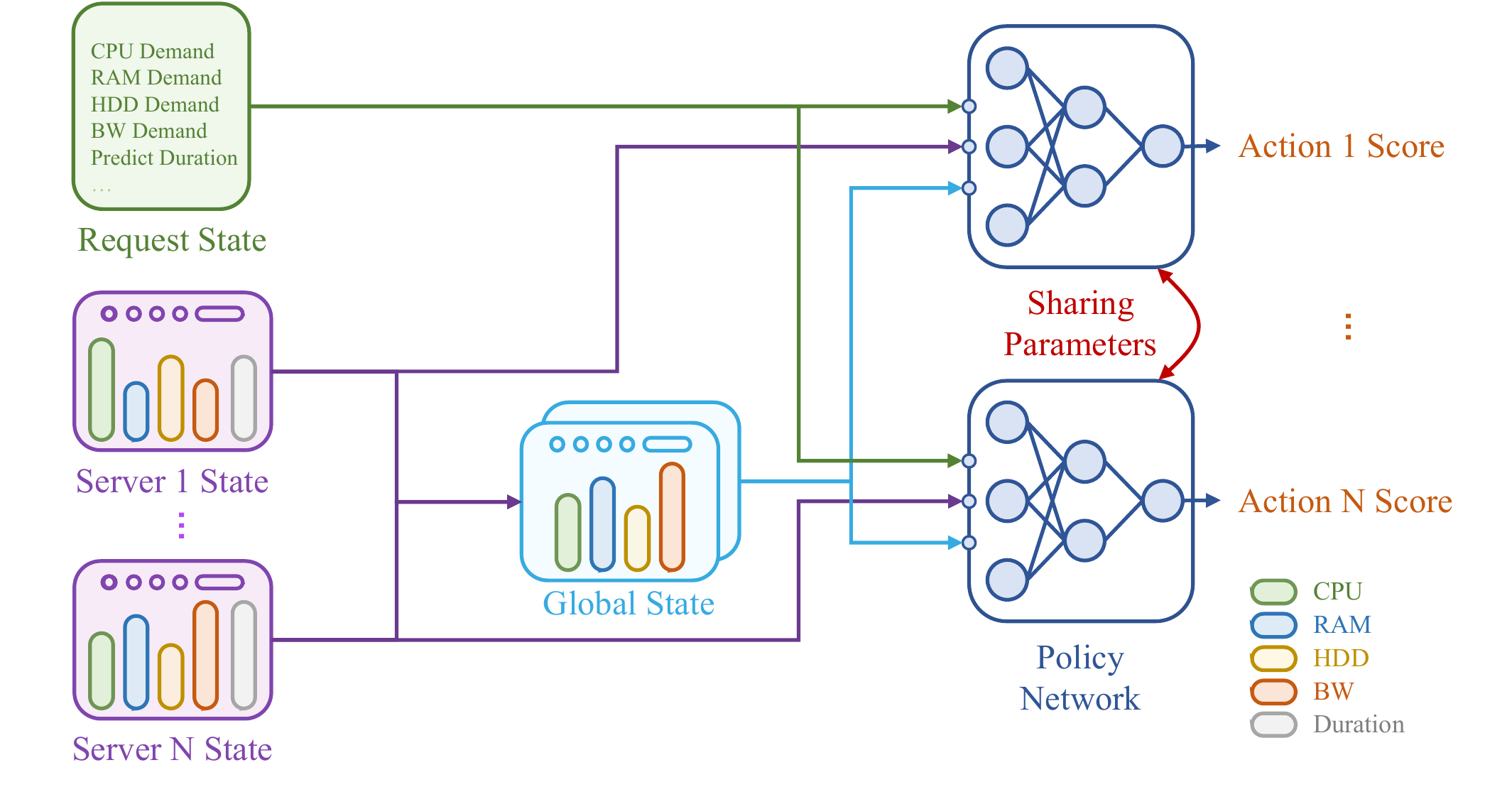}
  \caption{Inference process of the scalable policy network.}
  \label{fig: scalable policy}
\end{figure}

As shown in Fig.\ref{fig: scalable policy}, the policy network achieves the scalability with the global state management and the sharing parameters among $N$ copies, each of which corresponds to one available server. The input of this policy network is divided into three components: the request state, the individual server state, and the global state. The global state refers to the mean and the standard deviation among all individual server states, either of which is basically a $40$-dimensional vector with 4 resources looking forward for $10$ timesteps and $1$ remaining duration. There are $N$ fully connected networks jointly for the decision making. All networks share the same weight parameters and can receive the shared global state of servers as well as the $5$-dimensional request state. In addition, each $i$-th network will individually receive the $41$-dimensional states of the $i$-th server. For the decision making, each $i$-th network outputs a score. A higher score indicates better load balance and predicted idle time reduction. At last, the action with the highest score will be finally selected and the corresponding server will be chosen to construct the connection. By this architecture, the whole policy is able to scale with the number of servers and can intuitively be generalized to different data services. The update of the parameter-sharing networks is simple. Each network targets a server. No matter how main servers there are, their corresponding networks share the same weights. And the weights are represented as an individual in the evolutionary multi-objective training framework.

\subsection{Evolutionary multi-objective training framework}
As discussed above, the underlying load balancer has two objectives (see Eqs.\eqref{equ: balance fitness}-\eqref{equ: idleness fitness}), i.e., the long term rewards. In traditional RL policy training, it is common to aggregate these two rewards into a single one by weights sum. Unfortunately, due to the different scales of those two objectives, it is quite difficult to set proper weights, and the two objectives might not easily be satisfied. 

In recent years, evolutionary algorithms based RL policy training methods have been warmly studied, and form a new topic called evolutionary RL \cite{qian2021derivative, bai2023evolutionary,jianye2022erl}. This type of new methods enjoys several advantages over the policy-gradient methods which use gradient descent to optimize the policy. Apart from the ability of global search \cite{yang2021parallel} and learning from sparse reward \cite{wang2021evolutionary}, one major benefit of evolutionary RL is the flexibility of incorporating the evolutionary multi-objective framework \cite{Bindong2015survey} to train the policy. Inspired by this progress, we propose to employ the evolutionary multi-objective algorithms \cite{QIAN2023241, wang2023regularity} to optimize the weights of the policy directly for three reasons. First, evolutionary multi-objective algorithms have been the mainstream methods for multi-objective optimization problems and have successfully shown their power in many real-world applications \cite{hong2020efficient, liu2021effective, liu2022reliable,libingdong2023}. Second, it does not need to set the aggregation weights manually but directly compare solutions with the domination relation. Third, it suffices to output a set of optimal solutions called non-dominated Pareto set, which is equivalent to the optima with different weights sum aggregation\cite{chen2021interactive,coello2020ind}, thus offering the users the flexibility to select the policy online. Here we employ the well-established NSGA-II \cite{deb2002a}, i.e., almost the most successful evolutionary multi-objective algorithm developed by Deb et al. in 2002, to construct the evolutionary multi-objective training framework for the policy. For more details on NSGA-II, please refer to \cite{deb2002a}. It is worth to mention that several advanced evolutionary multi-objective reinforcement learning methods have been proposed recently \cite{shen2020generating,song2022evolutionary,xu2020prediction,abels2019dynamic,van2014multi}. Though they are not developed for the underlying routing problem, their algorithmic insights can shed light on the further improvement of training the proposed policy network.

\begin{figure}[hbt]
  \centering
  \includegraphics[width=0.95\linewidth]{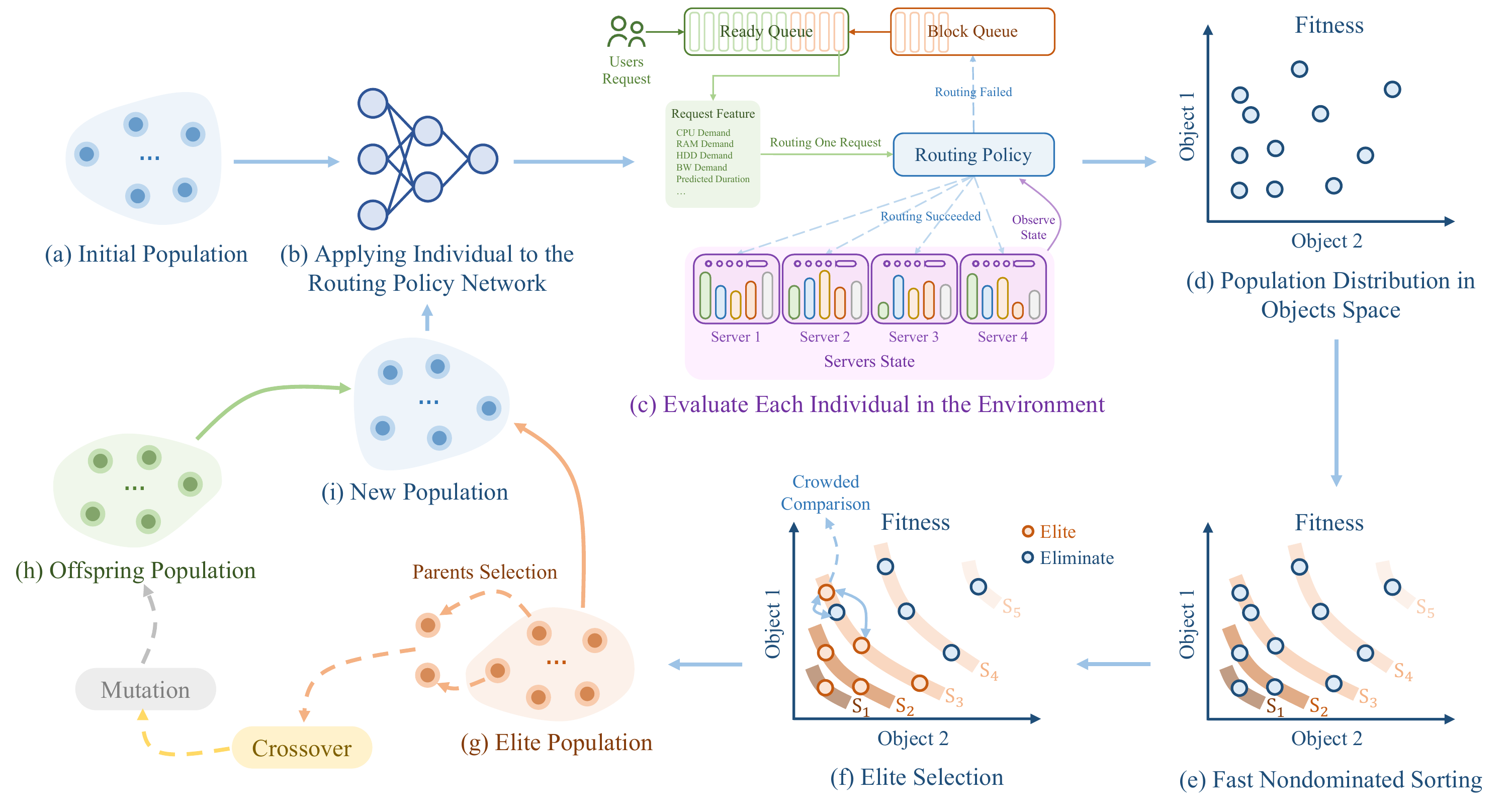}
  \caption{Overview of the Evolutionary Multi-objective Training Framework.}
  \label{fig: MERL-LB}
\end{figure}

The proposed NSGA-II based training framework shown in Fig.\ref{fig: MERL-LB} is an iterative process. At the beginning of the training (step (a)), a population of the routing policy network is randomly initialized. Second, we evaluate each individual policy in the simulation environment (steps (b)-(c)). Here, the objective functions in Eqs.\eqref{equ: balance fitness}-\eqref{equ: idleness fitness} are directly used for evaluation, as evolutionary algorithms usually measure the accumulated performance of a candidate policy over the whole simulation period \cite{wang2021evolutionary}. It is important to note that the settings of the simulation environment, e.g., the requests sequence and the servers’ initial states, are identical for the evaluation of all individual policies. After the evaluation, we obtain the bi-objective fitness values for all individuals in the current population, which distributes in a $2$-dimensional objective space (step (d)). In this space, the closer to the origin, the better the individual will be. 

Then, the individuals in the current population are sorted by the fast non-dominated sorting of NSGA-II, and $L$ non-dominated solution sets $S_1,S_2,\ldots,S_L$  are obtained (step (e)). Here, a non-dominated solution set means that an individual within the solution set cannot dominate any other individuals within the same solution set, but is dominated by the sets prior to it, e.g., $S_1$ dominates $S_2,\ \ldots,\ S_L$. More specifically, an individual A dominates an individual B means that A is not inferior to B in all objectives and is superior to B in at least one objective. The elite individuals are selected from the current population based on the precedence of the non-dominated solution sets and the crowding comparison (step (f)), where the crowding comparison algorithm is used to avoid selecting similar individuals and thus ensure the diversity of the population. Finally, based on the elite population, the offspring population is generated with typical operators of evolutionary algorithms (step (h)) and forms. It forms the new population for the next iteration together with the elite population (steps (g)-(i)). And then the iteration continues until some stop-criteria is met, and the individuals in $S_1$ are output as the trained policies.

\begin{figure}[hbpt]
  \centering
  \includegraphics[width=0.55\linewidth]{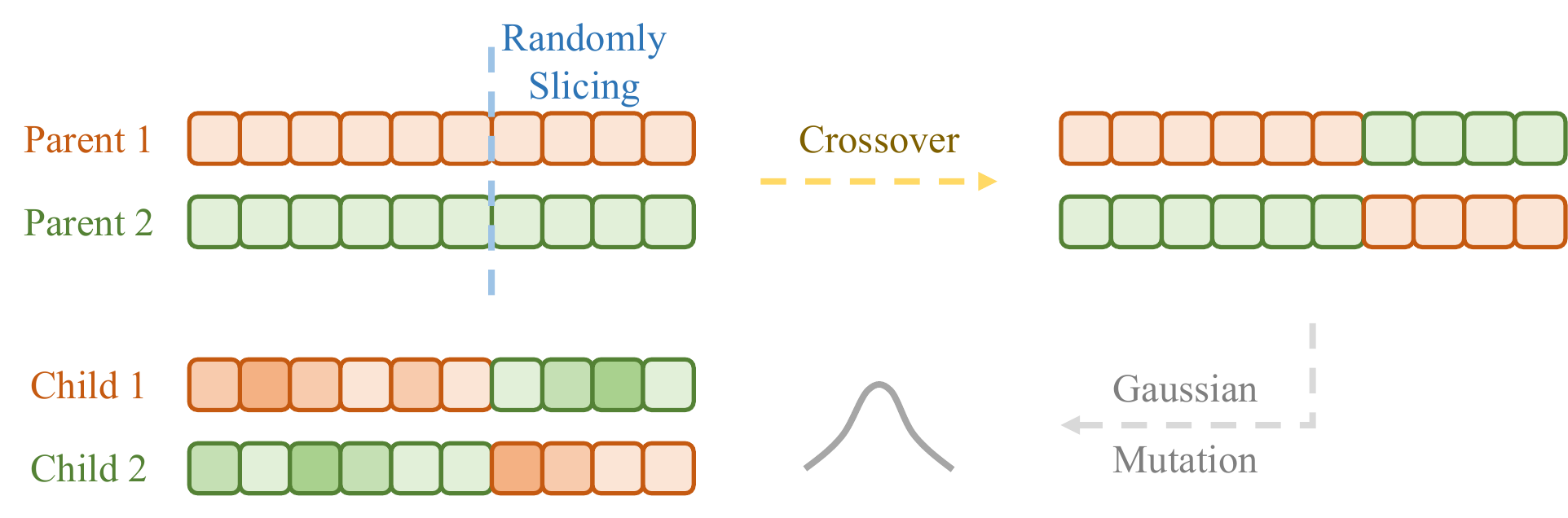}
  \caption{The process of generating offspring.}
  \label{fig: generating offspring}
\end{figure}

In the above framework, each policy is represented as a vector of weight parameters of the network described in section \ref{section: scalable policy}. The offspring generation step is to generate two new vectors based on two parent vectors. As shown in Fig.\ref{fig: generating offspring}, the two parent vectors crossover at a uniformly randomly chosen point to produce two tentative vectors. Then, based on the two tentative vectors, two offspring vectors are generated by the commonly used Gaussian mutation operator. Specifically, for each $i$-th weight parameter of the policy network, with a probability of $\gamma$, it will be mutated by sampling from the Gaussian distribution $N\left(\theta_i,\beta\right)$, where $\theta_i$ denotes the $i$-th weight parameter of a tentative vector, and $\beta$ denotes the magnitude of the mutation.

\subsection{Actions mask operator}
Note that, while simulating the individual policies within the step (c) of Fig.\ref{fig: MERL-LB}, the policies will not be updated unless the simulation ends. This might be a major difference between evolutionary RL and traditional policy gradient based RL. Though it has been revealed that this evolutionary RL training paradigm has merits for those scarce reward problems \cite{kaushik2018multi}, it indeed causes an issue in the load balancer. That is, the policy may be trapped in the situation of continuously routing the incoming requests to the same server, making that server fully utilized quickly. As a result, the subsequent requests cannot be connected to that server and thus have to be put into the block queue. Once the block queue overflows, the simulation will be stopped, and the evaluated objective fitness of that policy will be very poor.  

As the initial policies are randomly generated, they are highly likely to face this issue. And the training may easily fail. To stabilize the training process, a simple action mask operator is proposed. That is, after $N$ sharing networks in a policy outputting their scores, those servers whothat cannot satisfy the hardware demand of the current request will be mask and the score of the corresponding sharing networks will be set to 0. After that, the server with the highest score will be selected. By doing so, the above issue has been addressed and the action mask operator can be viewed as a specific kind of diversity enhancement.

\section{Experiments}
The experimental studies are designed to answer the following research questions.

(1) How does MERL-LB perform on the task of reducing idleness without disconnecting user connections, in comparisons with traditional methods?

(2) How does the proposed evolutionary multi-objective framework facilitate the training of RL-based routing policy, in terms of the convergence rate and the diversified user options?

(3) How is the stability of MERL-LB against different loads?

(4) How is the scalability of the proposed parameter-sharing routing policy against different numbers of user requests and servers?

(5) How does MERL-LB tolerate to the different variances of the user connection time?


\subsection{Simulation environment}
The experiment simulates a commonly encountered scenario of financial data services. The simulation starts at the opening of the stock market and ends after all users disconnect from the servers naturally, which will be later than the close of the stock market. For the major experiments, we simulate around $1500$ user requests coming up gradually to $10$ servers. The coming up duration simulates between the opening and close time of the stock market, which is set to $2$ hours. Later, for the stability and scalability tests, the number of user requests varied up to $7500$, and the number of available servers increases from $10$ to $50$.

All simulations are conducted on a Discrete Event Simulation based virtual data center implemented with Python\footnote{https://github.com/zlaom/MERL-LB}. The configurations of the virtual data center are shown in Table \ref{tab: simulated virtual data center}. The number of servers in the data center is set to $10$; Each server has four resources (CPU, RAM, HDD, and BW), and the capacity of each resource is set to $500$ units; At the beginning of the simulation, all the resource utilization is initialized to zero. The queue sizes of the ready queue and block queue in the data center are fixed to $200$ connections, respectively. The event of the simulation process is triggered five times every minute. That is, the minimum simulation interval is set to $12$ seconds. The predicted range for the future states of the data center is $120$ minutes, among which $h=10$ timestamps are sampled to represent the prediction of the system's future state. 

\begin{table}[htbp]
\centering
\begin{minipage}[t]{0.51\linewidth}
\centering
\makeatletter\def\@captype{table}
\caption{Configurations of the simulated virtual data center.}
\label{tab: simulated virtual data center}
\begin{tabular}{ll}
    \toprule
    Parameters         & Value        \\
    \midrule
    time\_step         & 12 (seconds)  \\
    server\_num        & 10           \\
    cpu\_capacity      & 500 (units)   \\
    ram\_capacity      & 500 (units)   \\
    hdd\_capacity      & 500 (units)   \\
    bw\_capacity       & 500 (units)   \\
    ready\_queue\_size & 200          \\
    block\_queue\_size & 200          \\
    predicted range    & 120 (minutes) \\
    future\_sample     & 10           \\
    \bottomrule
\end{tabular}
\end{minipage}
\begin{minipage}[t]{0.48\linewidth}
\centering
\makeatletter\def\@captype{table}
\caption{Simulation data generation parameter setting.}
\label{tab: simulation data generation}
\begin{tabular}{ll}
    \toprule
    Parameters          & Value        \\
    \midrule
    data\_time          & 120 (minutes) \\
    time\_step          & 12 (seconds)  \\
    mean\_req\_num      & 3            \\
    min\_res\_req       & 0 (units)     \\
    max\_res\_req       & 10 (units)    \\
    min\_user\_duration & 1 (minutes)   \\
    max\_user\_duration & 120 (minutes) \\
    \bottomrule
\end{tabular}
\end{minipage}
\end{table}


\subsection{Compared algorithms}
Four classic rule-based heuristic routing methods and two RL based algorithms were used for comparisons. As discussed in section \ref{sec: related works}, three industrial algorithms, like Random routing, Round Robin, and the Least Connection are selected as they merely consider load balancing. Another algorithm named Least Duration Gap is designed based on Least Connection to only consider the reduction of idleness, ignoring the load balancing. The core idea of the Least Duration Gap is to greedily route the user requests with similar predicted remaining duration to the same server with the following rule:
\begin{equation}
    a=\arg\min_{i}\left(\left|d_{s}^{i}-d_{r}\right|\right),
\end{equation}
wher$e d_s^i$ indicates the maximum predicted remaining duration of the $i$-th server, while $d_r$ is the predicted duration of the request. 

For the two RL based methods, they are designed to show the effectiveness of the proposed evolutionary multi-objective training framework. For this purpose, the Proximal Policy Optimization (PPO) \cite{schulman2017proximal} and the Independent-input Policy Gradient (IPG) \cite{mao2019variance} are employed to train the proposed parameter-sharing policy. The former is a well-established RL method that has been successfully applied in many fields. The latter is a variant of the original IPG method for job scheduling problems that are similar to the traditional load balancing problem. Both of them follow the policy gradient training framework. PPO calculates the gradient based on the temporal-difference error by an incomplete sequence of interactions, while IPG calculates the gradient with a complete sequence of interactions. The immediate rewards of a routing action at the specific $t$-th timestep can be simply derived from Eqs.\eqref{equ: balance fitness}-\eqref{equ: idleness fitness} by eliminating the summation over $t$, as shown in Eqs.\eqref{equ: blance reward}-\eqref{equ: idle reward}. And the two rewards are linearly aggregated with fixed weights to form the training signal, as shown in Eqs.\eqref{equ: reward}.
\begin{equation}
    \text{reward}_{\text {balance}}^{t}=\frac{1}{4} \sum_{r=1}^{4} \sqrt{\frac{\sum_{i=1}^{N}\left(x_{r i}^{t}-\mu_{r}^{t}\right)^{2}}{N}},
    \label{equ: blance reward}
\end{equation}
\begin{equation}
    \text{reward}_{\text{idle}}^{t}=\frac{1}{N} \sum_{i=1}^{N} d_{i}^{t},
    \label{equ: idle reward}
\end{equation}
\begin{equation}
    \text{reward}^t=w_i\cdot\text{reward}_{\text {balance}}^{t}+w_2\cdot\text{reward}_{\text{idle}}^{t}.
    \label{equ: reward}
\end{equation}

\subsection{Implementation details}
The architecture of the parameter-sharing neural network is introduced as follows. The network is basically a multi-layer perceptron with a $126$-dimensional input layers, a $32$-dimensional hidden layers, and a 1-dimensional output layer. The Relu is used as the activation function of the hidden layer. The $126$-dimensional input refers to the $5$-dimensional resource state of each incoming request, the $41$-dimensional individual server state of the corresponding server, and the 80-dimensional global state with half for the average and half for the standard deviation of all individual servers on the $4$ types of hardware indicators along $10$ steps forward. The $1$-dimensional output is the score of routing the incoming request to the corresponding server. The network with the highest output score among all $N$ networks is found, and the corresponding server is chosen to be routed. The number of weight parameters of the network is 4097, and all $N$ networks share the same weights \footnote{The codes can be found https://github.com/zlaom/MERL-LB}. All three RL-based methods train this network through the experiments.



The major hyperparameters of the PPO-LB and IPG-LB are described as follows. For the major simulation settings, both PPO-LB and IPG-LB share the same parameters. The maximum number of training simulations is set to $750000$ for both algorithms, which is kept the same for MERL-LB. The learning rate of the network is set to $0.001$, the discount factor $\gamma$ is $0.99$, and the aggregation weights of the load balancing reward and the idle time reward are $0.5$, respectively, for both algorithms. The network of PPO-LB is updated every $256$ steps of interaction. Each interaction sequence is reused for $5$ epochs. The clipping threshold $\epsilon$ of PPO-LB is set to $0.2$. The weight $\alpha_1$ of the loss of the value network in the loss function is $0.5$, and the weight $\alpha_2$ of the entropy term coefficient is $0.01$. The batch size during training is set to $512$ for IPG-LB, and each data instance randomly samples $10$ interaction sequences using the same policy. In general, those parameters are all suggested by the original paper of PPO \cite{schulman2017proximal} and IPG \cite{mao2019variance}. The $4$ rule-based heuristics also do not involve hyperparameters.

For the major hyperparameters of the proposed MERL-LB, we set the population size to $50$, the number of elites selected from the population per iteration to $25$, and the number of offspring generated by elites to $25$. The probability of mutation on each variable is $0.25$, and the magnitude of mutation is $0.05$. The time budget of the simulation is the same as PPO-LB and IPG-LB for fairness. 

\begin{table}[htbp]
\centering
\caption{Performance of different algorithms for two objectives on a validation set of 10 servers with a 75\% load.}
\label{tab: overall performance of different algorithms}
\begin{tabular}{lcc}
\toprule
        Methods & $F_{\text{balance}}(\text{units})\downarrow$  & $F_{\text{idle}}(\text{minutes})\downarrow$  \\
\midrule
        Random & 30.15±2.55 & 106.26±0.83  \\ 
        Round Robin & 22.15±1.96 & 107.12±0.85  \\ 
        Least Connection & \textbf{16.27}±1.04 & 106.98±0.95  \\ 
        Least Duration Gap & 66.89±6.12 & \textbf{69.91}±1.56  \\ 
\midrule
        IPG-LB & 17.05±1.12 & \textbf{79.75}±1.41 \\ 
        PPO-LB & \textbf{16.23}±1.14 & 80.67±1.34 \\ 
\midrule
        MERL-LB-1  & \textbf{15.70}±1.17 & 88.68±1.56  \\ 
        MERL-LB 4  & 16.88±1.06 & 76.75±1.25  \\ 
        MERL-LB-6  & 23.42±1.50 & 72.44±1.26  \\ 
        MERL-LB-11  & 66.58±4.93 & 61.63±1.26  \\ 
        MERL-LB-25 & 178.77±4.88 & \textbf{46.48}±1.73 \\ 
\bottomrule
\end{tabular}
\end{table}

\subsection{Overall comparisons}
Table \ref{tab: overall performance of different algorithms} reports the overall comparison results between MERL-LB and the compared algorithms, and is divided into three parts. The top part shows the performance of the heuristic-based algorithms, the middle part is about the two policy gradient based reinforcement learning load balancers , and the bottom part shows representative solutions generated by the proposed MERL-LB algorithm, where the suffix indicates different individuals in the last population. 

First of all, the three RL-based methods generally perform better than the heuristic-based methods, especially on the load balancing objective. Furthermore, MERL-LB has been able to find the best solution for either objective and also offers different options for various balances between the two objectives.

It can be seen that MERL-LB-1 focuses more on the load balancing objective and performs the best among all algorithms. Compared with the Least Connection method, which merely targets such an objective and performs the best among the compared heuristics, MERL-LB-1 reduces the idle time by 17.11\% while achieving better load balancing performance. MERL-LB-25 focuses more on the idle time objective and can reduce up to 33.51\% over the Least Duration Gap method, who is specifically designed for reducing idleness. 

\begin{figure}[ht]
  \centering
  \includegraphics[width=0.65\linewidth]{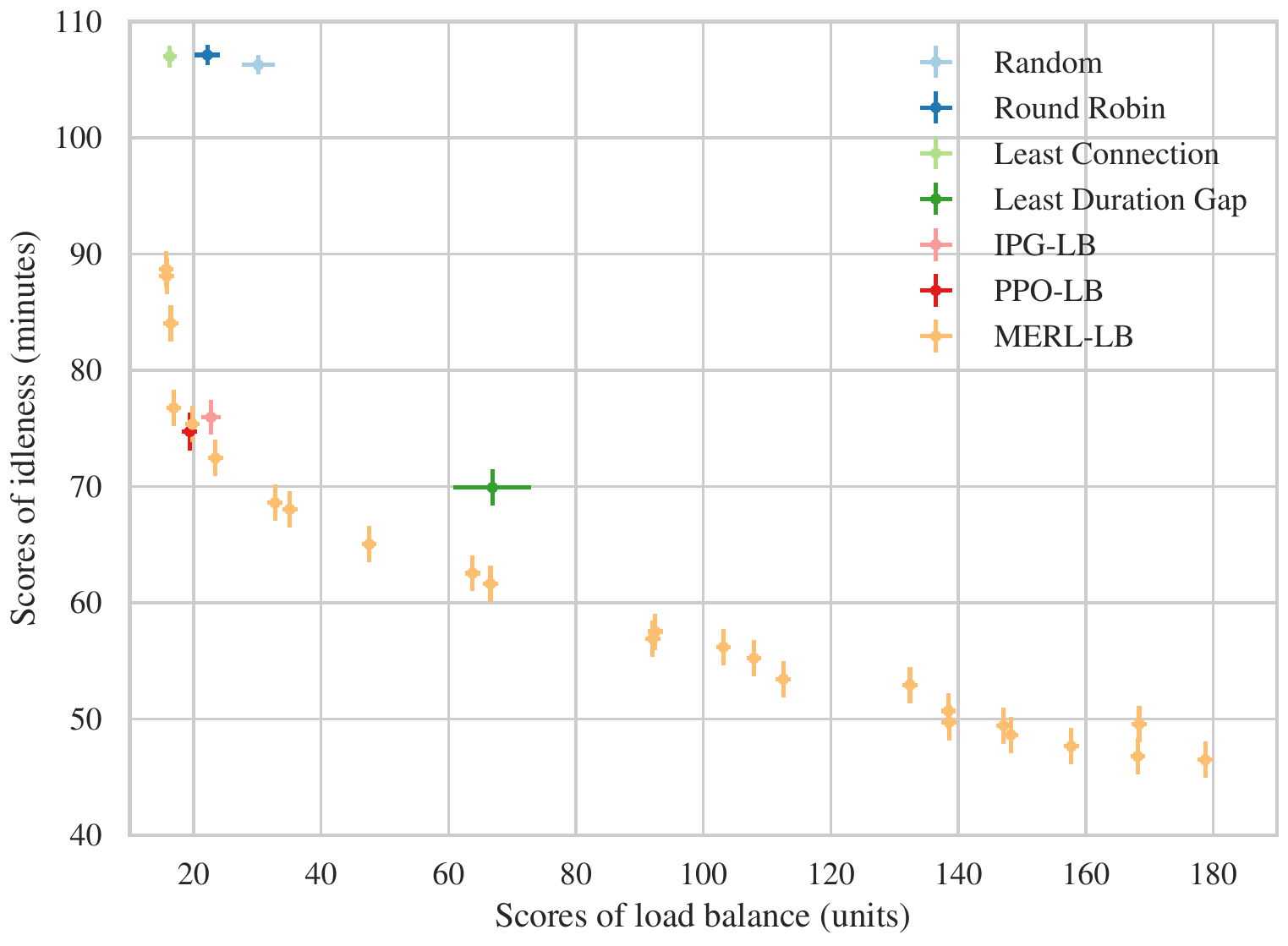}
  \caption{Performance of different algorithms on two objectives. The cross-like error bars indicate the standard deviation on both objectives.}
  \label{fig: overall performance of different algorithms}
\end{figure}

\begin{figure}[ht]
  \centering
  \includegraphics[width=0.5\linewidth]{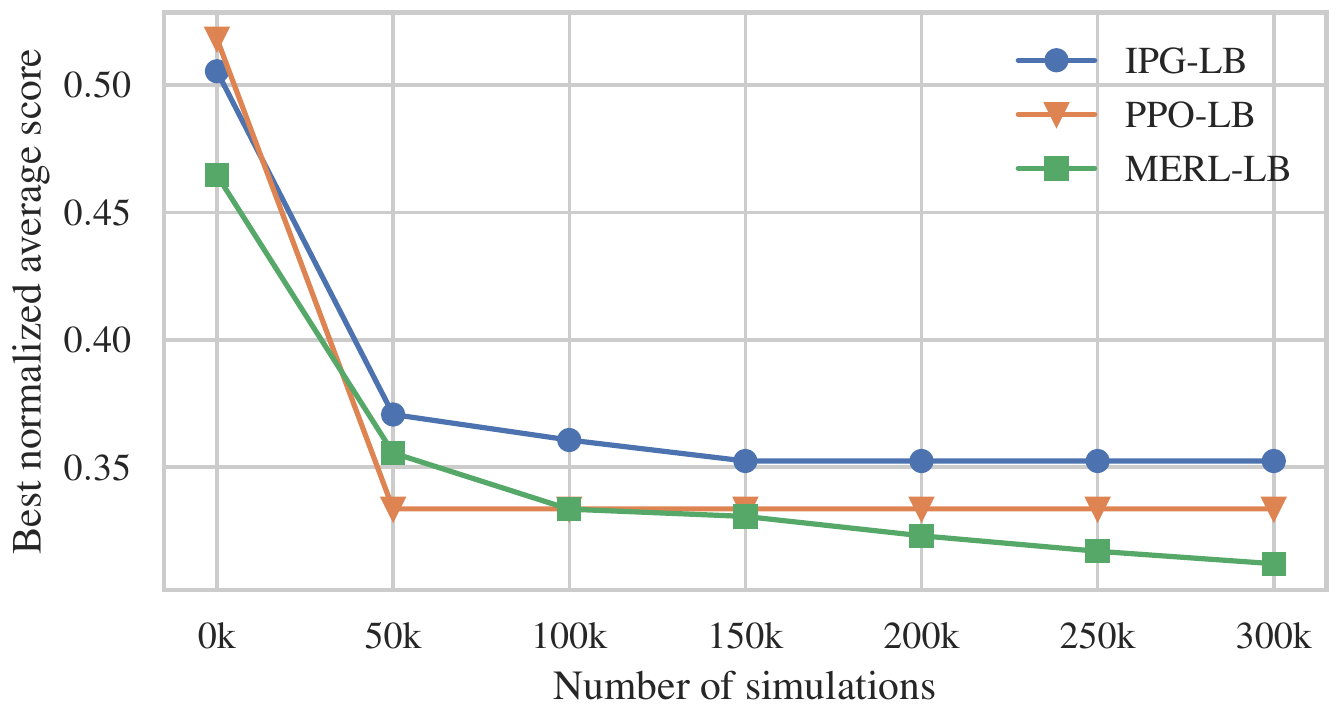}
  \caption{The convergence curves of RL-based load balancers on two objectives in the training process.}
  \label{fig: convergence curves of RL-based load balancers}
\end{figure}

The other 3 solutions of MERL-LB successfully show how evolutionary multi-objective optimization can offer users different options. MERL-LB-4 not only achieves a better load balancing objective than IPG-LB but reduces the idle time by 3.76\%. MERL-LB-6 achieves similar idleness to the Least Duration Gap method and can improve the load balancing objective by 64.99\%. MERL-LB-11 performs similarly to the Least Duration Gap in terms of the load balance while successfully reducing the idleness by 11.87\%. To summarize, the Pareto optima output by MERL-LB can effectively balance the two objectives automatically.

Fig.\ref{fig: overall performance of different algorithms} shows the distribution of the optimal policies generated by different algorithms in the objective space. The figure clearly shows that compared with PPO-LB and IPG-LB, MERL-LB can generate a set of diverse policies that are non-dominated to each other on both objectives. That means any one of the policies is a good choice for a specific need. By looking at the error bars of the policies in both directions (i.e., both objectives) in the figure shows that the policies generated by MERL-LB are mostly stable, especially compared with the Least Duration Gap method. Also note that, MERL-LB generalizes well as the testing instances are different from the training instances, while the Pareto policies trained by MERL-LB still distribute well on the objective space of the testing instances.

To further demonstrate the effectiveness of the proposed evolutionary multi-objective training framework, Fig.\ref{fig: convergence curves of RL-based load balancers} depicts how the three RL-based methods converge during the first 300000 simulations in the training process. There are two part of the training overhead, i.e., the algorithmic part and the simulation part. We repeat the training processes of IPG-LB, PPO-LB, and MERL-LB for 50 runs and the average of the computational time of each algorithm is recorded. It shows that the total computational costs at each iteration (one complete simulation) of IPG-LB, PPO-LB, and MERL-LB are 14.64 seconds, 9.3498 seconds, and 11.3127 seconds, indicating that MERL-LB is competitive in terms of the wall-clock training time\footnote{Experiments run on a workstation with 2 Intel(R) Xeon(R) Gold 6240 CPU (2.60GHz), in total with 72 threads, 377GB memory, and 4 TITANX GPUs.}. Furthermore, it has been found that the main overhead of training lies in the simulation process, as it costs 9.3426 seconds for simulating 1 thousand users. Therefore, we mainly use the number of simulations to measure the training cost of the algorithm in this paper. Each curve is composed of the scores over the training simulation. For each score, it is an average of two min-max normalized objective values. In short, all three methods have converged, and MERL-LB achieves better convergence rate. One major reason is that, as the optimization process goes, the conflict between two objectives becomes more significant. This makes PPO-LB and IPG-LB more difficult to express the non-dominated relationship between solutions with the weighted sum method. Simpler objective may impact the weighted objective more heavily and thus leads to significant bias to the search. As a result, PPO-LB and IPG-LB bias to the load balance objective and eventually converge slower with the indicator of normalized average score in Fig.\ref{fig: convergence curves of RL-based load balancers}.

\subsection{Impacts on different numbers of user requests}
\begin{figure*} [ht]
\centering
\subfloat[The impacts of different loads on the load balance objective.]{
    \includegraphics[width=0.8\linewidth]{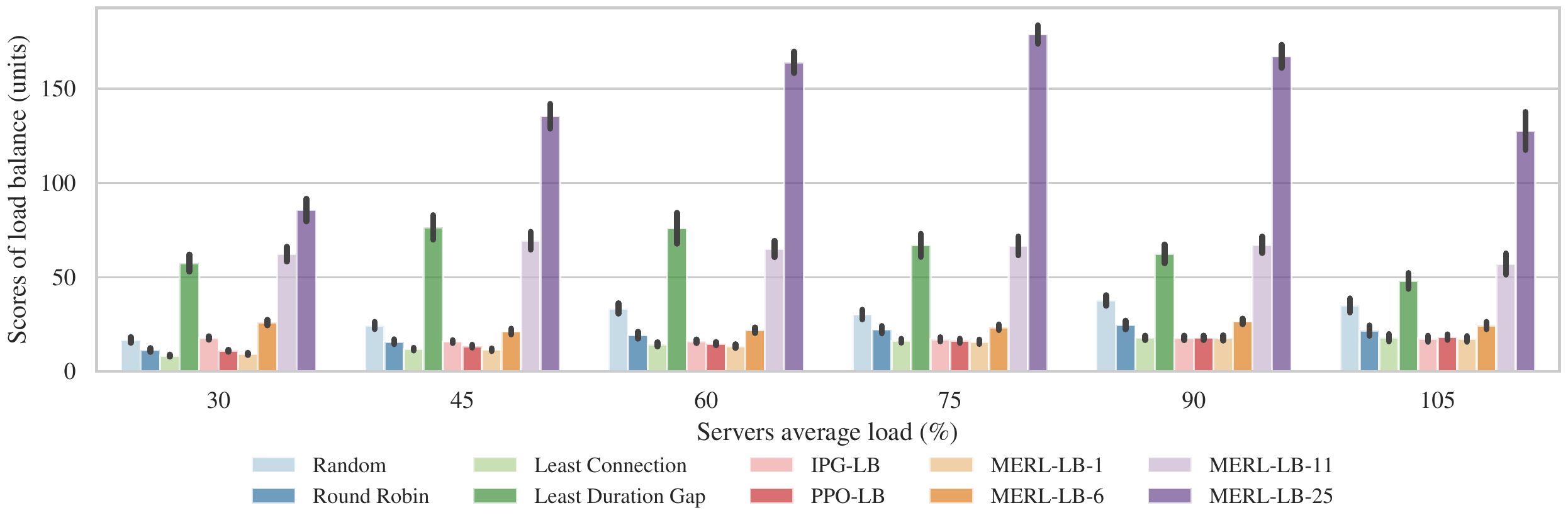}
}
\\
\subfloat[The impacts of different loads on the idleness objective.]{
    \includegraphics[width=0.8\linewidth]{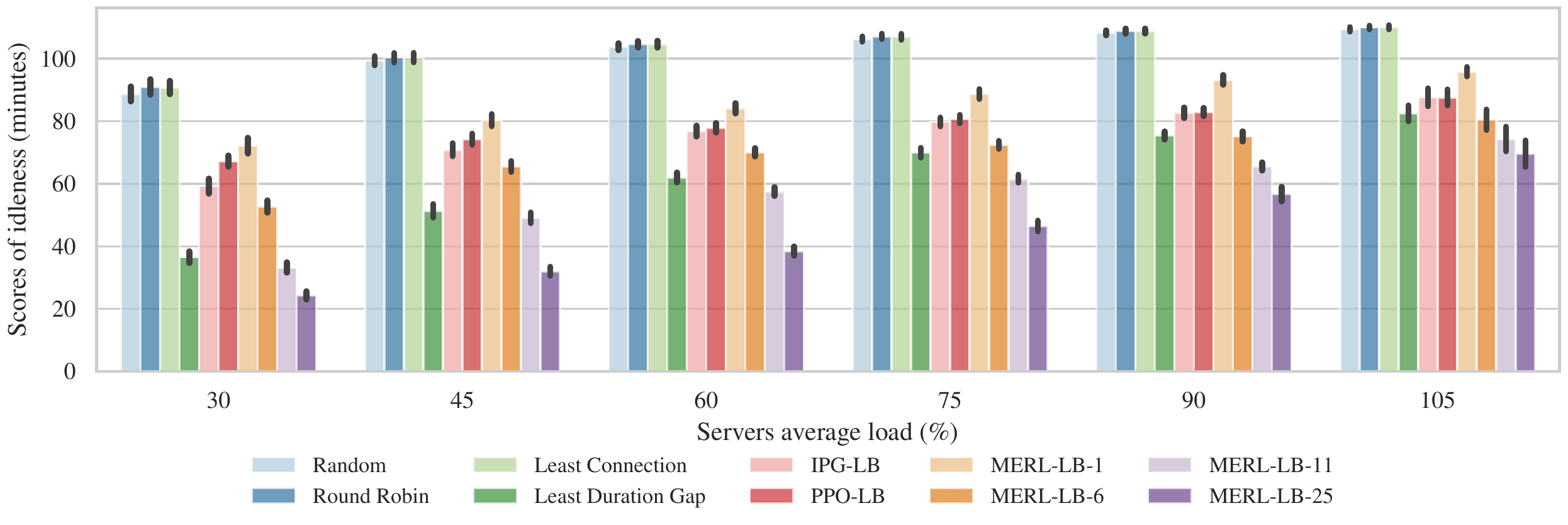} 
}
\caption{Performance of algorithms on two objectives under different loads, i.e., different numbers of user requests.}
\label{fig: performance of algorithms on two objectives under different loads} 
\end{figure*}

Fig.\ref{fig: performance of algorithms on two objectives under different loads} shows how different numbers of user requests may impact the performance of the algorithms on the two objectives. Through this group of experiments, the number of available servers is fixed to 10. The number of user requests is simulated from 600 to 2100 by increasing the mean value of the employed Poisson distribution. With a fixed 2 hours of the simulated opening of the market, not only does the total number of user requests increase but the concurrently incoming user requests at each time interval are enlarged, which leads to the increased loads of servers. As a result, this group of experiments assesses the algorithms for different numbers of user requests and different loads of servers in the same way. 

For the load balancing objective, the performance of Random, Round Robin, the Least Connection, PPO-LB, and MERL-LB-6 gradually deteriorate as the load increases. This is mainly because higher resource utilization rates may lead to a greater load imbalance among servers, as the standard deviation among servers can be larger. With the increase of loads, the load balancing performance of LDG, MERL-LB-11 and MERL-LB-25 all first deteriorated and then gradually improved. This is mainly because those algorithms all pay more attention to reducing idleness. When the load is at a low level, it leads to a greater imbalance among servers. However, when the load increases, even servers with lower loads will be assigned more requests, which will gradually reduce the differences in resource utilization rates among servers. 

For the idleness objective, as the load increases, the idle time of all algorithms increases accordingly. This is mainly because an increase in the load means an increase in the number of long connection requests, which would need more servers to serve the connections, while the number of servers is fixed at 10.

In general, MERL-LB-1 performs similarly to the Least Connection method at all loads and is able to reduce the idle time by 10-20\%. MERL-LB-6 achieves a similar idleness objective value to LDG at 75\%-105\% loads while it improves the load balancing objective by about 50\%-60\%. Compared to PPO-LB and IPG-LB, MERL-LB can generate diverse policies and consistently outperform the two at different loads. Therefore, MERL-LB performs more robustly overall when the number of requests or loads varies.

\subsection{Impacts on different numbers of available servers}
\begin{figure*} [htbp]
\centering
\subfloat[The impacts of different numbers of user requests on the load balance objective.]{
    \includegraphics[width=0.8\linewidth]{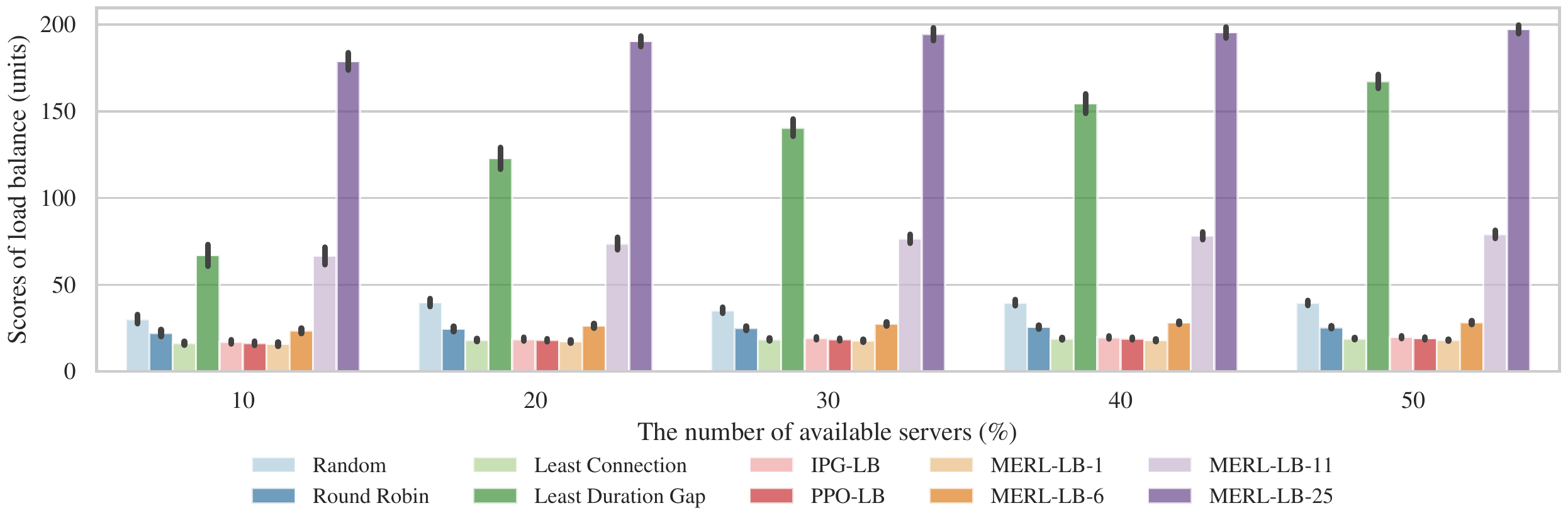}
}
\\
\subfloat[The impacts of different numbers of user requests on the idleness objective.]{
    \includegraphics[width=0.8\linewidth]{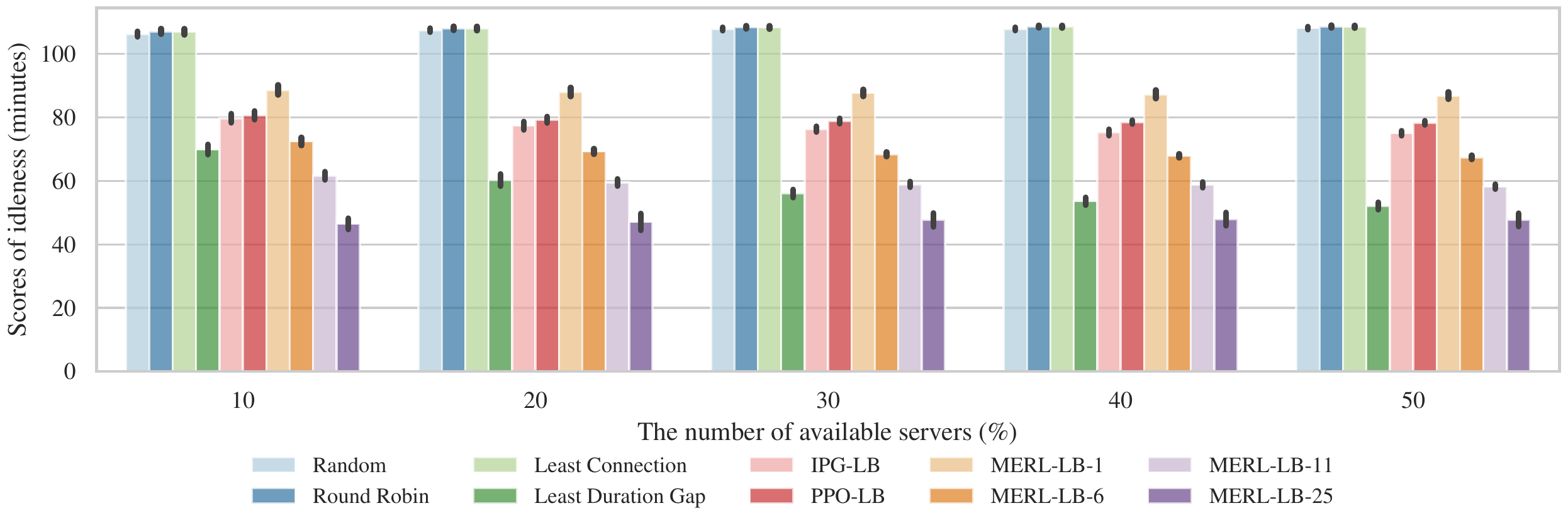} 
}
\caption{Performance of algorithms on two objectives with different numbers of servers.}
\label{fig: performance of algorithms on two objectives with different numbers of servers} 
\end{figure*}

Fig.\ref{fig: performance of algorithms on two objectives with different numbers of servers} shows the performance of the algorithms on two objectives under different numbers of servers, while the average load is fixed at 75\%. Thus, by varying the number of servers from 10 to 50, the number of user requests also changes proportionally, up to around 7500 requests in total for the case of 50 servers.  For the load balancing objective, it can be observed that as the number of servers increases, the load balance of each algorithm does not deteriorate too much, except the Least Duration Gap method, who is the only algorithm that never considers the load balance objective. Similarly, for the idleness objective, the performance of the Least Duration Gap method improves most significantly as the number of servers increased, though it still cannot outperform MERL-LB-25 in all tested cases. The other algorithms perform rather stably, regardless of the changed number of servers. Overall, this suggests that the proposed scalable policy network can be flexibly applied to scenarios with different numbers of servers and can maintain stable routing capabilities on the two objectives.

\subsection{Impacts on the predictability of user duration}
This work assumes that we can predict the duration of a user by statistically analyzing their connection time over a recent period (for example, the average value of the user's recent connection duration). In real cases, the actual duration of users may vary over time, and is highly unlikely identical to the predicted value. Thus, this group of experiments aims to assess how different degrees of the predictability may impact the performance of the algorithms, reflecting how effective MERL-LB can behave in real-world noisy environment.

For this purpose, we generate new testing data sequences based on the original 50 testing instances by adding a random noise to each user’s connection time. More specifically, the random noise follows a Gaussian distribution $N(\mu,\sigma)$, where $\mu$ is the duration of the original data, and $\sigma$ represents the standard deviation of noise, resulting in different predictability of the user connection time. The random noise is truncated at a $3\sigma$ level. 

\begin{figure*}[ht]
\centering
\subfloat[The impacts of different user predictability on the load balance objective.]{
    \includegraphics[width=0.8\linewidth]{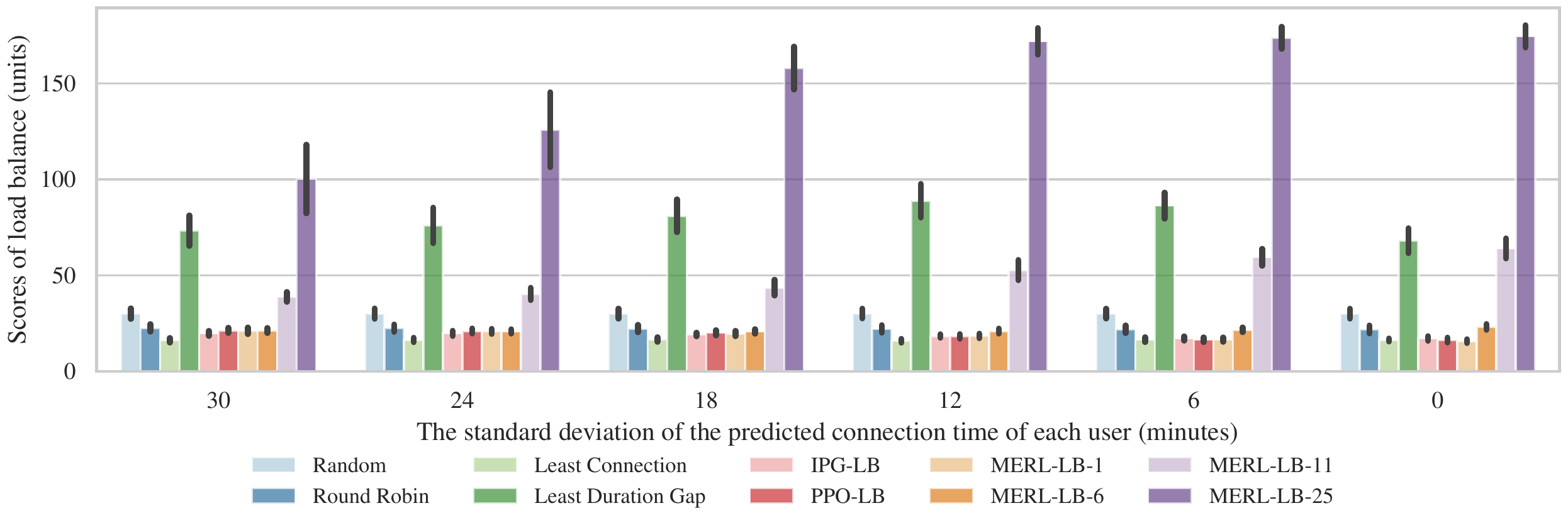}
}
\\
\subfloat[The impacts of different user predictability on the idleness objective.]{
    \includegraphics[width=0.8\linewidth]{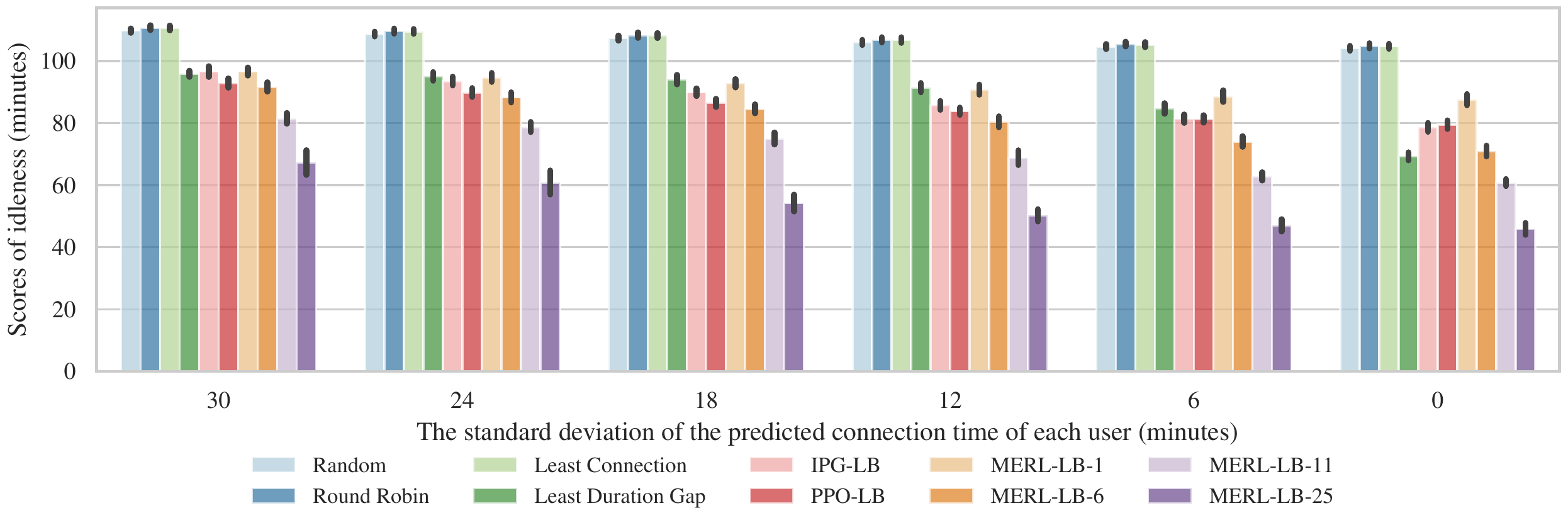} 
}
\caption{Performance of algorithms on two objectives under different user predictability.}
\label{fig: performance of algorithms on two objectives under different user predictability} 
\end{figure*}

Fig.\ref{fig: performance of algorithms on two objectives under different user predictability} shows the impacts of different user duration predictability on the algorithms for two objectives respectively. In general, for the three heuristics that do not consider the idleness objective, their performance does not improve when the predictability increases. For the Least Duration Gap method, which focuses on the idleness objective but does not rely on the prediction of user connections, its performance on the idleness objective improves near 20 minutes. For the other RL-based policies that share the same scalable routing network, the prediction of the user connections has been involved in their training processes. As can be seen, the various predictability does not deteriorate their performance unexpectedly. That is, as the predictability increases, their performance on the idleness objective becomes better similar to the Least Duration Gap method. This shows that with up to a standard deviation of 30 minutes error on the prediction of the user connection time (maximally 120 minutes), the proposed scalable routing network works quite stable.  

On the other hand, the policies that tend to focus more on the idleness objective deteriorate on the load balance objective as the predictability increases. The reason is that as the predictability of the user connection increases, the randomness of the data as well as the routing decreases. And the randomness usually forms a source of load balancing, as suggested by the Random method.

\subsection{Understanding the trade-off between two objectives}
To understand how the load balancing objective and the idleness objective are balanced during the learning of MERL-LB, we show how the two objectives change over time for the Least Duration Gap method, the Least Connection method, and MERL-LB-6. For the Least Duration Gap method, only the idleness objective is considered, while for the Least Connection method, only the load balancing objective is considered. MERL-LB-6 is selected here as it balances the two objectives well. We depict the curves of the idle time and the CPU resource utilization in Fig.\ref{fig: comparison of LDG, MERL-LB, and LC characteristics}. For clarity, the simulated opening period is enlarged to a full range of 6 hours to make the observation easier. As shown in Fig.\ref{fig: comparison of LDG, MERL-LB, and LC characteristics} (a), (b), (c), each curve in the figure represents the change of the remaining duration of each server per minute (known from the ground truth of the testing data). In Fig.\ref{fig: comparison of LDG, MERL-LB, and LC characteristics} (d), (e), (f), each curve in the figure represents the change of CPU resource utilization of each server per minute.

\begin{figure*} [ht]
\centering
\subfloat[The remaining duration of each server in Least Duration Gap.]{
    \includegraphics[width=0.29\linewidth]{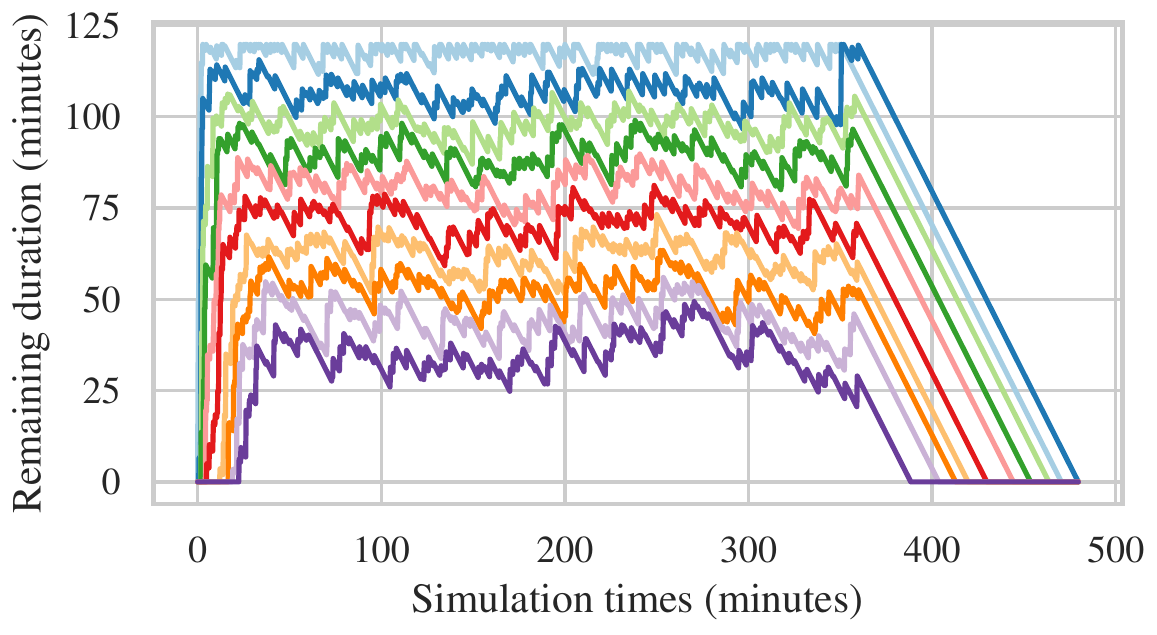}}\hspace{0.015\linewidth}
\subfloat[The remaining duration of each server in MERL-LB-6.]{
    \includegraphics[width=0.29\linewidth]{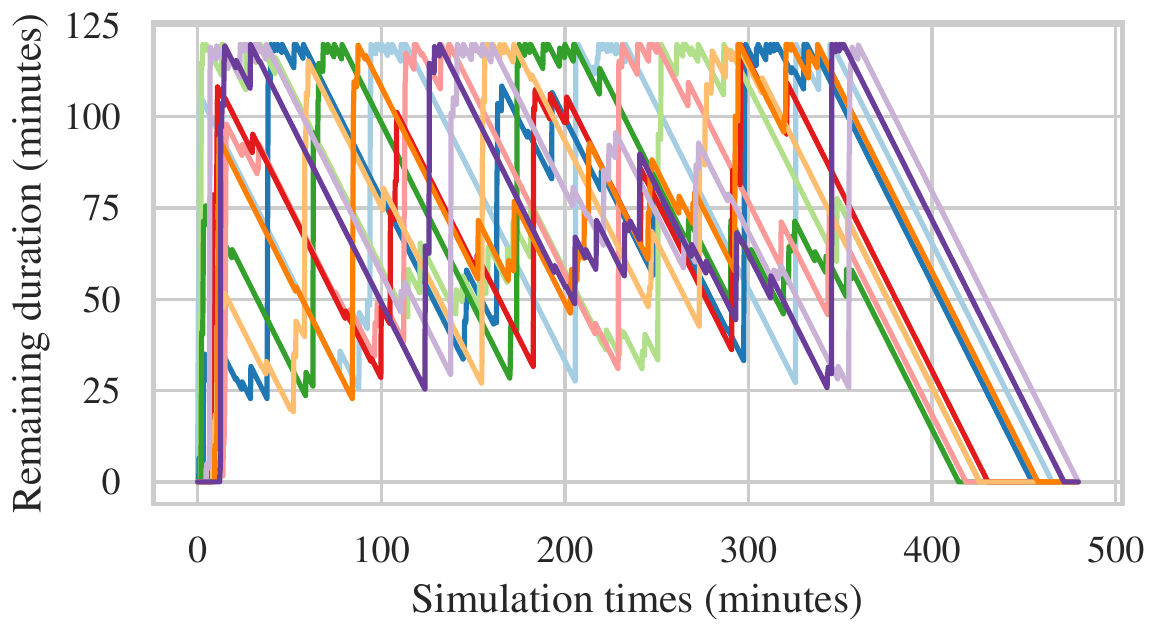}}\hspace{0.015\linewidth}
\subfloat[The remaining duration of each server in Least Connection.]{
    \includegraphics[width=0.29\linewidth]{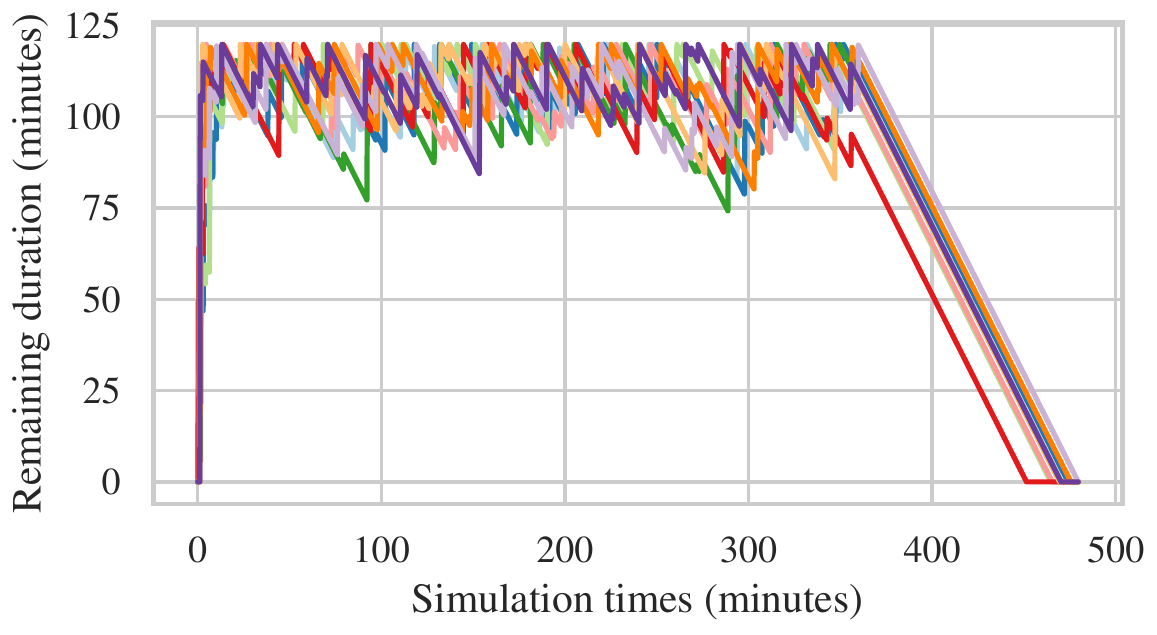}} \\
\subfloat[The CPU utilization of each server in Least Duration Gap.]{
    \includegraphics[width=0.29\linewidth]{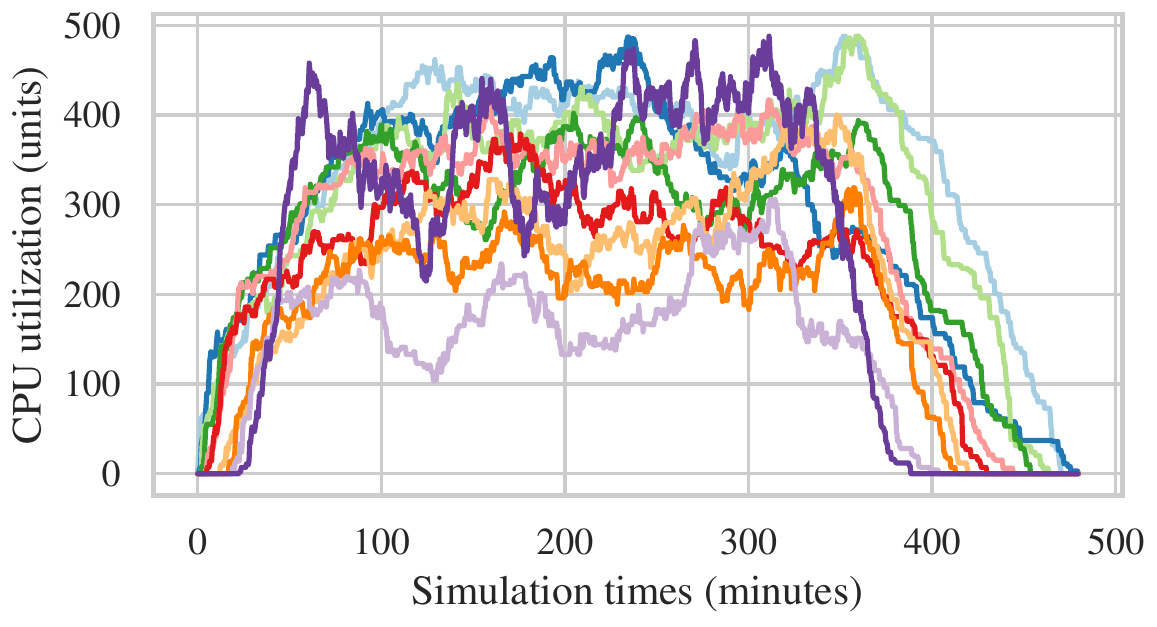}}\hspace{0.015\linewidth}
\subfloat[The CPU utilization of each server in MERL-LB-6.]{
    \includegraphics[width=0.29\linewidth]{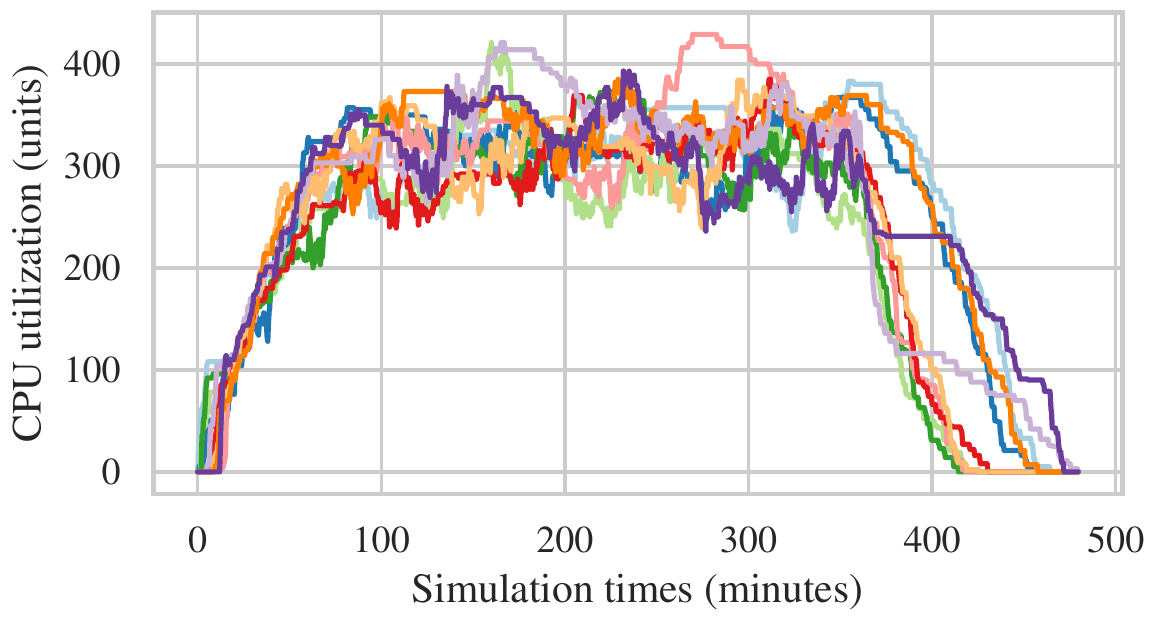}}\hspace{0.015\linewidth}
\subfloat[The CPU utilization of each server in Least Connection.]{
    \includegraphics[width=0.29\linewidth]{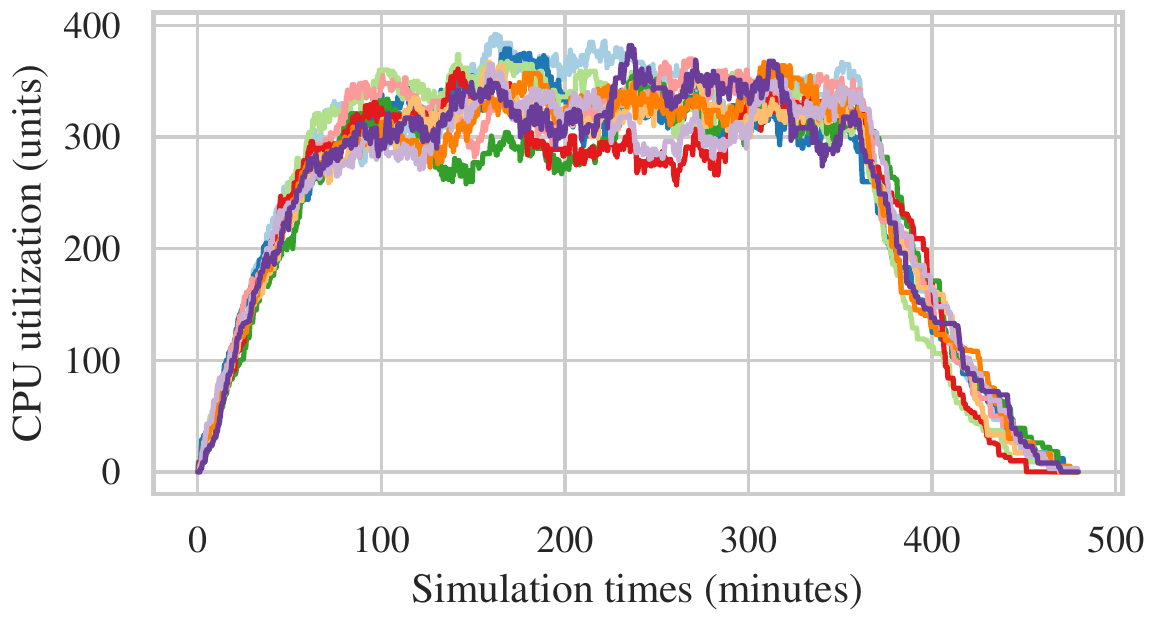}}
\caption{Comparison of Least Duration Gap, MERL-LB-6, and Least Connection characteristics on 10 servers.}
\label{fig: comparison of LDG, MERL-LB, and LC characteristics} 
\end{figure*}

During the opening period, the number of users continues to come up, so the CPU utilization of the 10 servers rises up and fluctuates, and becomes stable after the number of concurrent alive users is stabilized. Then, after the close of the market, no new user request comes up, and the number of connections as well as the CPU utilization, continues to decrease. The closer the utilization rate among servers per minute, the better the load balance is. In Fig.\ref{fig: comparison of LDG, MERL-LB, and LC characteristics} (e), each server's CPU resource utilization of MERL-LB-6 fluctuates about half of that in Fig.\ref{fig: comparison of LDG, MERL-LB, and LC characteristics} (d), so MERL-LB-6 has better load balancing performance than the Least Duration Gap. In Fig.\ref{fig: comparison of LDG, MERL-LB, and LC characteristics} (f), the Least Connection’s CPU utilization fluctuates even smaller than MERL-LB-6, so its load balancing performance is better correspondingly.

An interesting phenomenon can be observed by comparing Fig.\ref{fig: comparison of LDG, MERL-LB, and LC characteristics} (a), (b), (c) that the Least Duration Gap and MERL-LB-6 minimize the idleness very differently. In Fig.\ref{fig: comparison of LDG, MERL-LB, and LC characteristics} (a), the curve of the remaining duration of each server is completely separated and even shows a layered distribution. In this regard, the Least Duration Gap actually minimizes the idleness by clustering each incoming user request with all the alive connections based on their predicted remaining durations. On the other hand, as shown in Fig.\ref{fig: comparison of LDG, MERL-LB, and LC characteristics} (b), the curves of MERL-LB-6 overlap with each other and show a sawtooth pattern. This pattern may trade off the load balance better and contribute to better idleness as all curves drop to 0 in the shortest expectation time. In either above-mentioned case, the spectrum of the curves is much larger than that of the Least Connection, indicating that both the Least Duration Gap and MERL-LB-6 have a better trade-off on both objectives.

\subsection{Understanding the sawtooth pattern of MERL-LB}
\begin{figure*} [ht]
\centering
\subfloat[The ground truth remaining duration changes with CPU utilization over time.]{
    \includegraphics[width=0.4\linewidth]{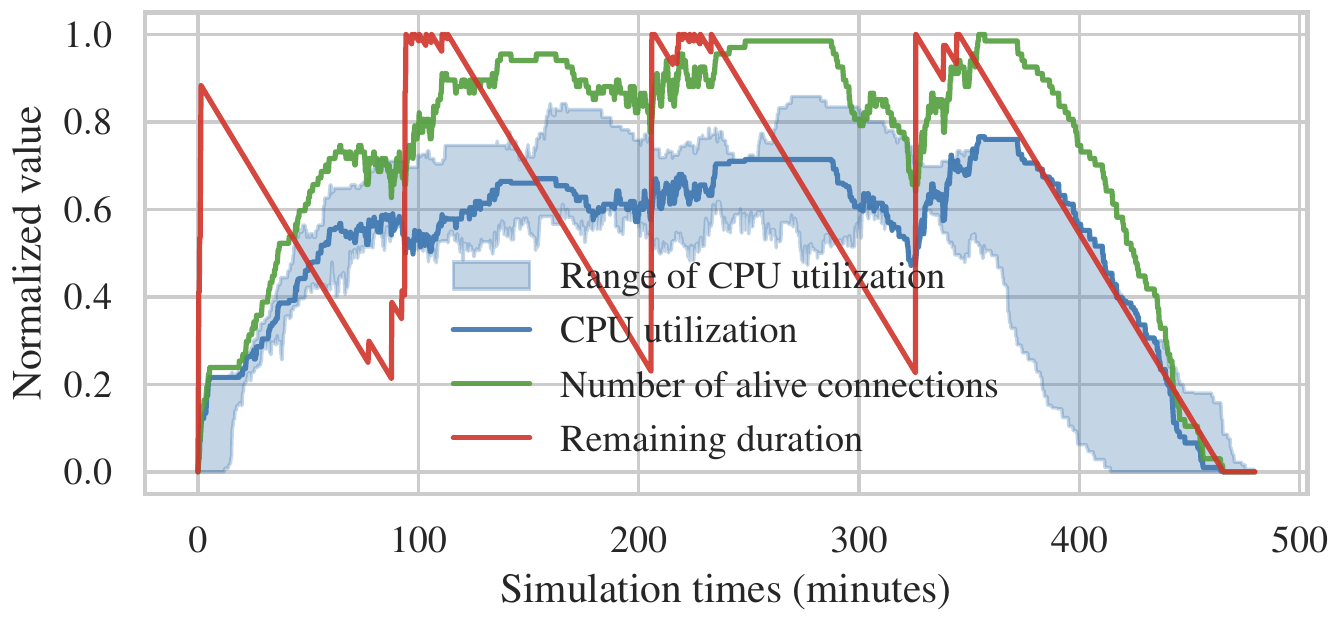}}\hspace{0.02\linewidth}
\subfloat[The ground truth remaining duration changes with RAM utilization over time.]{
    \includegraphics[width=0.4\linewidth]{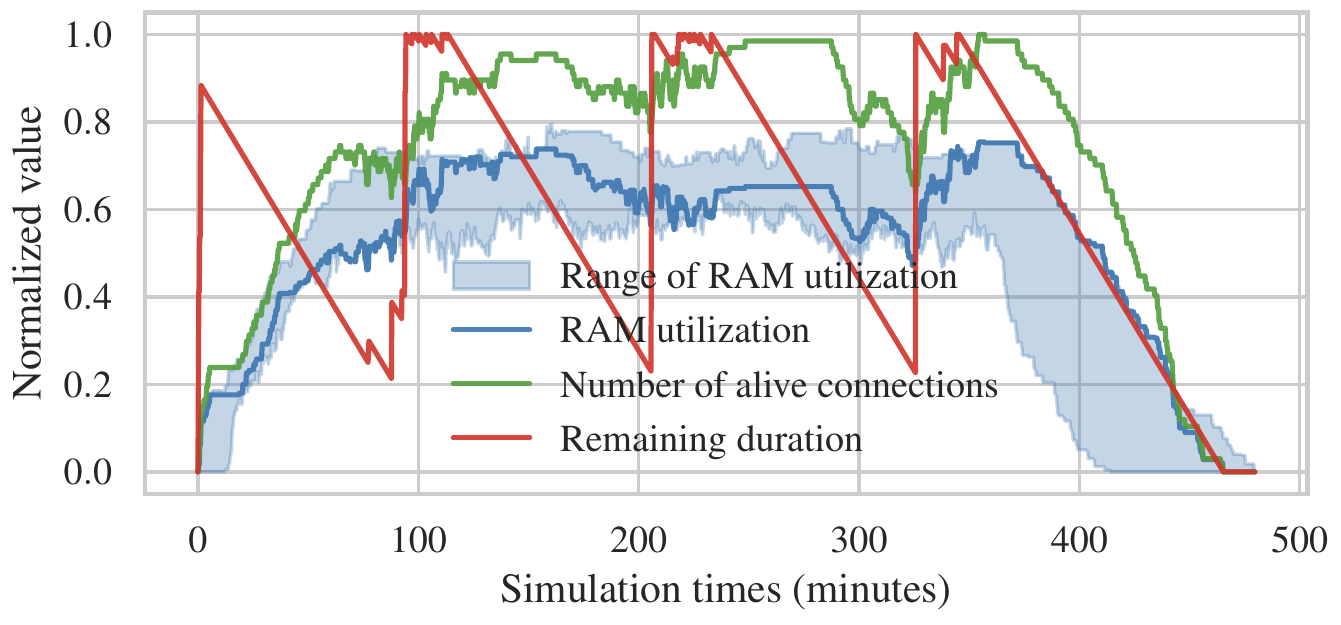}}\\
\subfloat[The ground truth remaining duration changes with HDD utilization over time.]{
    \includegraphics[width=0.4\linewidth]{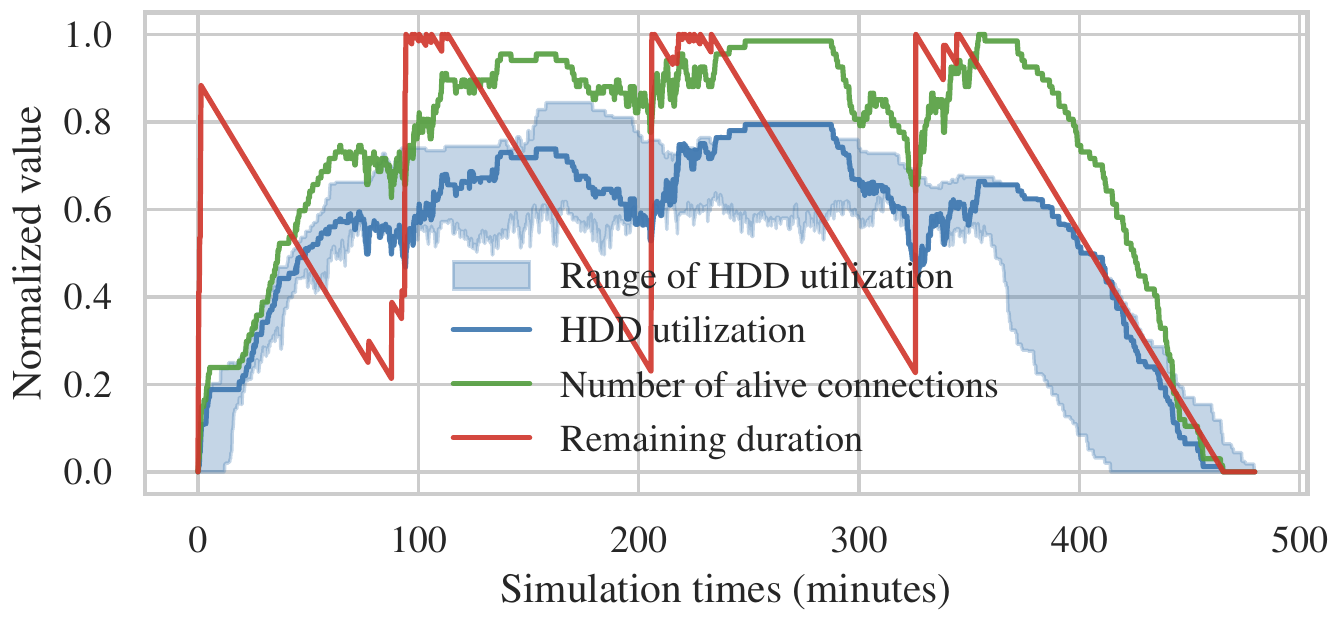}}\hspace{0.02\linewidth}
\subfloat[The ground truth remaining duration changes with BW utilization over time.]{
    \includegraphics[width=0.4\linewidth]{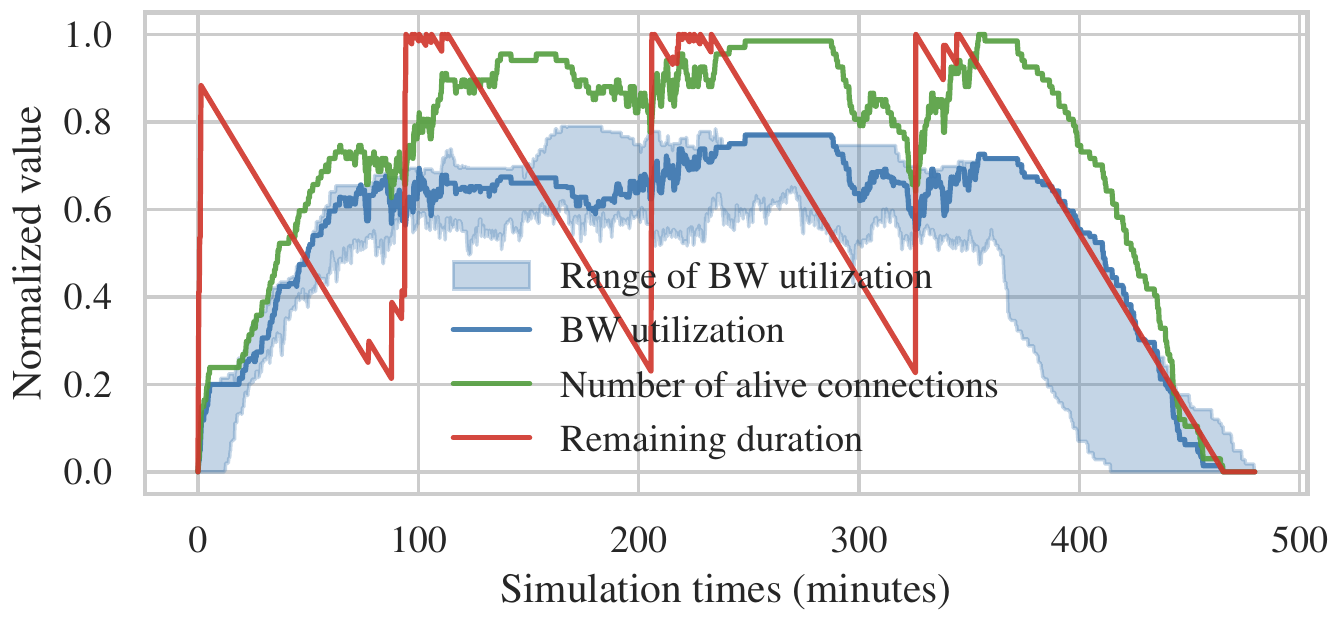}}
\caption{Analysis of the sawtooth pattern of MERL-LB-6 on a server.}
\label{fig: analysis of the sawtooth pattern of MERL-LB-6} 
\end{figure*}

To further understand the above sawtooth pattern of MERL-LB, the utilization rates of 4 types resources of MERL-LB-6 over time are visualized in Fig.\ref{fig: analysis of the sawtooth pattern of MERL-LB-6}. In each figure, 4 curves are depicted, i.e., the corresponding resource utilization of a server, the range of the corresponding resource utilization among all servers, the number of alive connections, and the ground truth remaining duration of this server.

It can be observed that the number of alive connections generally increases though keeps fluctuating. This indicates that new user requests are routed to the server continuously and some existing connections become disconnected coincidently. With the increase of the number of connections, the remaining duration does not go up accordingly. In fact, its sawtooth pattern suggests that the requests routed to this server are mostly with remaining durations smaller than that of this server. Thus, the remaining duration of this server gradually drops as time goes by. 

If we look at the place where the remaining duration of the server goes up significantly, it is usually found that the utilization of one resource type drops to the bottom of the corresponding range. If the utilization curve goes below its range, it means that the load balance in this resource type is enlarged. And it is necessary to route a request with a larger remaining duration to this server so that the resource utilization can be enlarged for a long time. On the other hand, once the utilization of one resource type goes to the top of the range, the number of alive connections will not increase. This is the opposite strategy learned by the policy to minimize the load balance by forcing the resource utilization not to exceed the range.

\subsection{A Real-world Application on the Alibaba Cluster-trace dataset}

The simulations have also been carried out in a real-world scenario with the well-known Alibaba Cluster-trace-v2017 dataset. This dataset was released by Alibaba Cloud in September 2017\footnote{https://github.com/alibaba/clusterdata}, consisting of detailed statistics of 11089 online service jobs and 12951 batch jobs co-locating on 1300 machines over 12 hours. The dataset has been found with four characteristics\cite{Chengzhi2017Alibaba}:1) Heterogeneous resource utilization across machines and workloads. 2) Greatly time-varying and multi-dimensional resource usages per workload and machine. 3) Imbalanced resource demands and runtime statistics (duration and task number) between online service and offline batch jobs. Thus, we believe this dataset is well suited to verify the proposed method. 

In this experiment, we randomly picked 500 thousands user requests with their hardware demands and duration from the online service jobs. The picked requests were used as one simulation scenario for training the routing policies. We randomly generated 50 testing simulation scenarios by disturbing the picked data. Thus, each of the 50 testing simulation scenarios also consisted of 500 thousands user requests that were different from the picked one. All the compared algorithms were involved in the comparison and the experimental settings kept the same to the synthetic simulation cases. 

\begin{figure}[ht]
  \centering
  \includegraphics[width=0.65\linewidth]{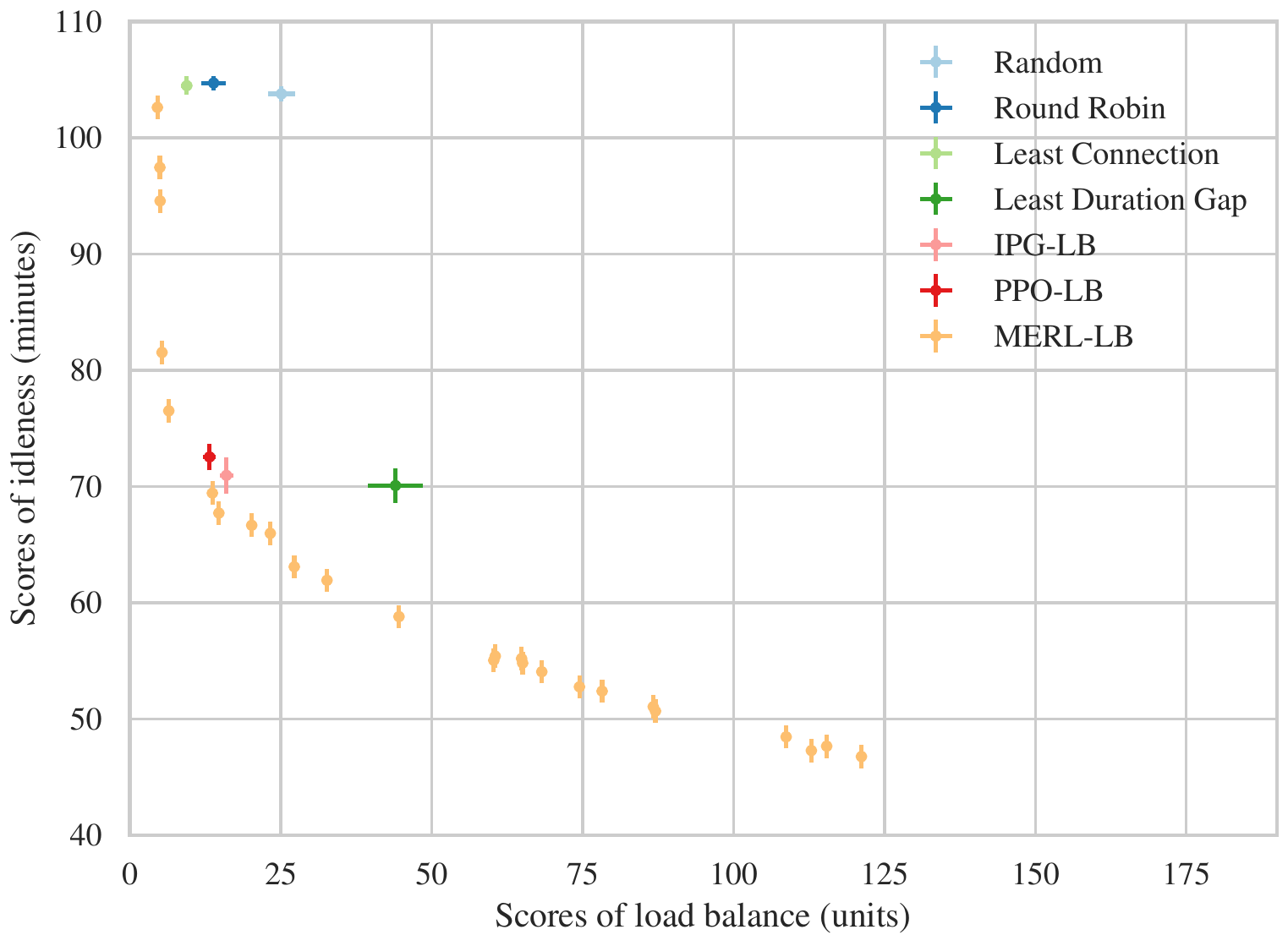}
  \caption{Performance of different algorithms on two objectives. The cross-like error bars indicate the standard deviation on both objectives.}
  \label{fig: overall performance of different algorithms on alibaba dataset}
\end{figure}

The results in Table \ref{tab: overall performance of different algorithms on alibaba} and Fig.\ref{fig: overall performance of different algorithms on alibaba dataset} clearly show that our MERL-LB can still beat the compared algorithms in the similar pattern with the synthetic scenarios. The general performance of each algorithm does not surprisingly change. First, the pareto solutions obtained by MERL-LB still distribute well. Second, there are always solutions of MERL-LB can significantly outperform the compared algorithms on either objective. Third, for each of the compared algorithm, there is always at least one solution of MERL-LB that dominates its output on both objectives. This verifies the effectiveness of the proposed MERL-LB in real-world datasets.

\begin{table}[htbp]
\centering
\caption{Solution distribution of different algorithms in the objective space.}
\label{tab: overall performance of different algorithms on alibaba}
\begin{tabular}{lcc}
\toprule
        Methods & $F_{\text{balance}}(\text{units})\downarrow$  & $F_{\text{idle}}(\text{minutes})\downarrow$  \\
\midrule
Random             & 25.11±2.23 & 103.77±0.66 \\
Round Robin        & 13.87±1.99 & 104.70±0.62 \\
Least Connection   & \textbf{9.43}±0.91  & 104.48±0.79   \\
Least Duration Gap & 43.99±4.50 & \textbf{70.08}±1.48  \\
\midrule
IPG-LB             & 15.99±1.09 & \textbf{70.95}±1.55  \\
PPO-LB             & \textbf{13.20}±1.06 & 72.54±1.16  \\
\midrule
MERL-LB-1  & \textbf{4.56}±0.39   & 102.61±1.01 \\
MERL-LB-5  & 6.45±0.54   & 76.51±1.74 \\
MERL-LB-6  & 13.67±1.29  & 69.43±0.99  \\
MERL-LB-12 & 44.58±4.28  & 58.80±1.93 \\
MERL-LB-25 & 121.19±6.73 & \textbf{46.76}±2.89\\
\bottomrule
\end{tabular}
\end{table}

\section{Conclusions}
This paper first introduced the new issue of reducing idleness in the financial servers where the user connections were only allowed to be naturally disconnected by users. This paper targeted the issue as a constrained online routing problem and identified the limitations of existing load balancers in not considering user connection duration information. To address this issue, this paper proposes to take into account the user connection duration information and model the problem as a bi-objective RL problem, i.e., the load balancing objective and the idleness objective. With this modeling, a parameter-sharing routing policy is designed for varied numbers of servers, and an evolutionary multi-objective training framework is proposed based on NSGA-II to train the policy. 

Extensive experimental studies have been conducted to reveal the advantages of the proposed method in detail. It is found that the proposed method generally beats comparable algorithms, including the traditional heuristics and advanced RL-based ones. The evolutionary multi-objective training framework not only facilitates policy training with a much faster convergence rate but also offers a set of diverse non-dominated policies for users’ options. To consolidate the experiments, we tested the algorithms with different numbers of user requests and different numbers of servers. It was observed that the proposed method performed in a quite stable manner due to the proposed scalable routing policy. Then, the proposed method generated a sawtooth pattern of decision-making strategy, which has been analyzed to be beneficial in balancing the two objectives. Finally, the proposed method is verified on a real-world dataset released by Alibaba Cloud. The comparisons again support the effectiveness of MERL-LB.

We look forward to applying this policy to real-world financial cloud systems soon. In this regard, it is important to study the more robust and efficient evolutionary training method for MERL-LB in a realistic noisy environment \cite{bian2021robustness} and expensive simulation environment \cite{hao2023enhancing}. Moreover, novel heuristics or solvers \cite{liu20213howgood} inspired by the effective sawtooth pattern can be studied with sound theoretic guarantees.




\Acknowledgements{This work was supported in part by National Natural Science Foundation of China (Grants 62272210, 62331014, and 62250710682), in part by the Guangdong Provincial Key Laboratory (Grant 2020B121201001), in part by the Program for Guangdong Introducing Innovative and Entrepreneurial Teams (Grant 2017ZT07X386), in part by the CCF-Tencent Open Fund (Grant RAGR20220110), and in part by the Market Data Cloud Joint laboratory between Shenzhen Securities Information Co., Ltd. and Southern University of Science and Technology (Grant XB11ZC20210189).}






\bibliographystyle{unsrt}
\bibliography{egbib}

\begin{thebibliography}{10}

\bibitem{Elasticity}
Yahya Al-Dhuraibi, Fawaz Paraiso, Nabil Djarallah, and Philippe Merle.
\newblock Elasticity in cloud computing: State of the art and research
  challenges.
\newblock {\em IEEE Transactions on Services Computing}, 11(2):430--447, 2018.

\bibitem{Qu2019Auto}
Chenhao Qu, Rodrigo~N. Calheiros, and Rajkumar Buyya.
\newblock Auto-scaling web applications in clouds: A taxonomy and survey.
\newblock {\em ACM Comput. Surv.}, 51(4):33, jul 2018.

\bibitem{qureshi2009cutting}
Asfandyar Qureshi, Rick Weber, Hari Balakrishnan, John Guttag, and Bruce Maggs.
\newblock Cutting the electric bill for internet-scale systems.
\newblock In {\em Proceedings of the ACM SIGCOMM 2009 conference on Data
  communication}, pages 123--134, 2009.

\bibitem{shastri2016transient}
Supreeth Shastri, Amr Rizk, and David Irwin.
\newblock Transient guarantees: Maximizing the value of idle cloud capacity.
\newblock In {\em SC'16: Proceedings of the International Conference for High
  Performance Computing, Networking, Storage and Analysis}, pages 992--1002.
  IEEE, 2016.

\bibitem{Stoll2006Elec}
Hans~R. Stoll.
\newblock Electronic trading in stock markets.
\newblock {\em Journal of Economic Perspectives}, 20(1):153--174, March 2006.

\bibitem{li2019cloud}
Feifei Li.
\newblock Cloud-native database systems at alibaba: Opportunities and
  challenges.
\newblock {\em Proceedings of the VLDB Endowment}, 12(12):2263--2272, 2019.

\bibitem{easley2011microstructure}
David Easley, M~Lopez De~Prado, and Maureen O’Hara.
\newblock The microstructure of the flash crash.
\newblock {\em Journal of Portfolio Management}, 37(2):118--128, 2011.

\bibitem{mirobi2019dynamic}
G~Justy Mirobi and L~Arockiam.
\newblock Dynamic load balancing approach for minimizing the response time
  using an enhanced throttled load balancer in cloud computing.
\newblock In {\em 2019 International Conference on Smart Systems and Inventive
  Technology (ICSSIT)}, pages 570--575. IEEE, 2019.

\bibitem{kansal2012cloud}
Nidhi~Jain Kansal and Inderveer Chana.
\newblock Cloud load balancing techniques: A step towards green computing.
\newblock {\em IJCSI International Journal of Computer Science Issues},
  9(1):238--246, 2012.

\bibitem{Kumar2019issues}
Pawan Kumar and Rakesh Kumar.
\newblock Issues and challenges of load balancing techniques in cloud
  computing: A survey.
\newblock 51(6):35, 2019.

\bibitem{Chengzhi2017Alibaba}
Chengzhi Lu, Kejiang Ye, Guoyao Xu, Cheng-Zhong Xu, and Tongxin Bai.
\newblock Imbalance in the cloud: An analysis on alibaba cluster trace.
\newblock In {\em 2017 IEEE International Conference on Big Data (Big Data)},
  pages 2884--2892, 2017.

\bibitem{shafiq2022load}
Dalia~Abdulkareem Shafiq, N.Z. Jhanjhi, and Azween Abdullah.
\newblock Load balancing techniques in cloud computing environment: A review.
\newblock {\em Journal of King Saud University-Computer and Information
  Sciences}, 34(7):3910--3933, 2022.

\bibitem{johora2022load}
Fatema~Tuj Johora, Iftakher Ahmed, Md~Ashiqul~Islam Shajal, and Rony Chowdhory.
\newblock A load balancing strategy for reducing data loss risk on cloud using
  remodified throttled algorithm.
\newblock {\em International Journal of Electrical and Computer Engineering},
  12(3):3217, 2022.

\bibitem{marinescu2017approach}
Dan~C Marinescu, Ashkan Paya, John~P Morrison, and Stephen Olariu.
\newblock An approach for scaling cloud resource management.
\newblock {\em Cluster Computing}, 20:909--924, 2017.

\bibitem{sim2003ant}
Kwang~Mong Sim and Weng~Hong Sun.
\newblock Ant colony optimization for routing and load-balancing: survey and
  new directions.
\newblock {\em IEEE transactions on systems, man, and cybernetics-Part A:
  systems and humans}, 33(5):560--572, 2003.

\bibitem{gures2022machine}
Emre Gures, Ibraheem Shayea, Mustafa Ergen, Marwan~Hadri Azmi, and Ayman~A
  El-Saleh.
\newblock Machine learning based load balancing algorithms in future
  heterogeneous networks: A survey.
\newblock {\em IEEE Access}, 10:37689--37717, 2022.

\bibitem{brar2016meta}
Gurveer~Kaur Brar and Amit Chhabra.
\newblock Meta-heuristics based load balancing algorithms in grid and clouds-a
  review.
\newblock In {\em 2016 International Conference on Electrical, Electronics, and
  Optimization Techniques (ICEEOT)}, pages 2938--2943. IEEE, 2016.

\bibitem{farag2021congestion}
Hossam Farag and {\v{C}}edomir Stefanovi{\v{c}}.
\newblock Congestion-aware routing in dynamic iot networks: A reinforcement
  learning approach.
\newblock In {\em 2021 IEEE Global Communications Conference (GLOBECOM)}, pages
  1--6. IEEE, 2021.

\bibitem{schulman2017proximal}
John Schulman, Filip Wolski, Prafulla Dhariwal, Alec Radford, and Oleg Klimov.
\newblock Proximal policy optimization algorithms.
\newblock {\em arXiv preprint arXiv:1707.06347}, 2017.

\bibitem{sajjan2017load}
RS~Sajjan and Biradar~Rekha Yashwantrao.
\newblock Load balancing and its algorithms in cloud computing: a survey.
\newblock {\em International Journal of Computer Sciences and Engineering},
  5(1):95--100, 2017.

\bibitem{alakeel2010guide}
Ali~M Alakeel et~al.
\newblock A guide to dynamic load balancing in distributed computer systems.
\newblock {\em International journal of computer science and information
  security}, 10(6):153--160, 2010.

\bibitem{chen2017clb}
Shang-Liang Chen, Yun-Yao Chen, and Suang-Hong Kuo.
\newblock Clb: A novel load balancing architecture and algorithm for cloud
  services.
\newblock {\em Computers \& Electrical Engineering}, 58:154--160, 2017.

\bibitem{afzal2019load}
Shahbaz Afzal and G~Kavitha.
\newblock Load balancing in cloud computing--a hierarchical taxonomical
  classification.
\newblock {\em Journal of Cloud Computing}, 8(1):22, 2019.

\bibitem{carrion2022kubernetes}
Carmen Carri{\'o}n.
\newblock Kubernetes scheduling: Taxonomy, ongoing issues and challenges.
\newblock {\em ACM Computing Surveys}, 55(7):1--37, 2022.

\bibitem{Kashani2023Load}
Mostafa~Haghi Kashani and Ebrahim Mahdipour.
\newblock Load balancing algorithms in fog computing.
\newblock {\em IEEE Transactions on Services Computing}, 16(2):1505--1521,
  2023.

\bibitem{trivella2016load}
Alessio Trivella and David Pisinger.
\newblock The load-balanced multi-dimensional bin-packing problem.
\newblock {\em Computers \& Operations Research}, 74:152--164, 2016.

\bibitem{basu2019learn}
Debabrota Basu, Xiayang Wang, Yang Hong, Haibo Chen, and St{\'e}phane Bressan.
\newblock Learn-as-you-go with megh: Efficient live migration of virtual
  machines.
\newblock {\em IEEE Transactions on Parallel and Distributed Systems},
  30(8):1786--1801, 2019.

\bibitem{zhu2022qos}
Jianyong Zhu, Renyu Yang, Xiaoyang Sun, Tianyu Wo, Chunming Hu, Hao Peng,
  Junqing Xiao, Albert~Y Zomaya, and Jie Xu.
\newblock Qos-aware co-scheduling for distributed long-running applications on
  shared clusters.
\newblock {\em IEEE Transactions on Parallel and Distributed Systems},
  33(12):4818--4834, 2022.

\bibitem{zhang2023smaf}
Shuai Zhang, Yunfei Guo, Zehua Guo, Hongchao Hu, and Guozhen Chen.
\newblock Smaf: a secure and makespan-aware framework for executing serverless
  workflows.
\newblock {\em Science China Information Sciences}, 66(3):139105, 2023.

\bibitem{wang2022solving}
Gai-Ge Wang, Da~Gao, and Witold Pedrycz.
\newblock Solving multiobjective fuzzy job-shop scheduling problem by a hybrid
  adaptive differential evolution algorithm.
\newblock {\em IEEE Transactions on Industrial Informatics}, 18(12):8519--8528,
  2022.

\bibitem{kandel2012enterprise}
Sean Kandel, Andreas Paepcke, Joseph~M Hellerstein, and Jeffrey Heer.
\newblock Enterprise data analysis and visualization: An interview study.
\newblock {\em IEEE transactions on visualization and computer graphics},
  18(12):2917--2926, 2012.

\bibitem{salimans2017evolution}
Tim Salimans, Jonathan Ho, Xi~Chen, Szymon Sidor, and Ilya Sutskever.
\newblock Evolution strategies as a scalable alternative to reinforcement
  learning.
\newblock {\em arXiv preprint arXiv: 1703.03864}, 2017.

\bibitem{qian2021derivative}
Hong Qian and Yang Yu.
\newblock Derivative-free reinforcement learning: a review.
\newblock {\em Frontiers of Computer Science}, 15(6):156336, 2021.

\bibitem{bai2023evolutionary}
Hui Bai, Ran Cheng, and Yaochu Jin.
\newblock Evolutionary reinforcement learning: A survey.
\newblock {\em Intelligent Computing}, 2:0025, 2023.

\bibitem{jianye2022erl}
HAO Jianye, Pengyi Li, Hongyao Tang, Yan Zheng, Xian Fu, and Zhaopeng Meng.
\newblock Erl-re $^{2}$: Efficient evolutionary reinforcement learning with
  shared state representation and individual policy representation.
\newblock In {\em The Eleventh International Conference on Learning
  Representations}, 2022.

\bibitem{yang2021parallel}
Peng Yang, Qi~Yang, Ke~Tang, and Xin Yao.
\newblock Parallel exploration via negatively correlated search.
\newblock {\em Frontiers of Computer Science}, 15:1--13, 2021.

\bibitem{wang2021evolutionary}
Yutong Wang, Ke~Xue, and Chao Qian.
\newblock Evolutionary diversity optimization with clustering-based selection
  for reinforcement learning.
\newblock In {\em International Conference on Learning Representations},
  virtual, 2022.

\bibitem{Bindong2015survey}
Bingdong Li, Jinlong Li, Ke~Tang, and Xin Yao.
\newblock Many-objective evolutionary algorithms: A survey.
\newblock {\em ACM Comput. Surv.}, 48(1), sep 2015.

\bibitem{QIAN2023241}
Chao Qian, Dan-Xuan Liu, Chao Feng, and Ke~Tang.
\newblock Multi-objective evolutionary algorithms are generally good:
  Maximizing monotone submodular functions over sequences.
\newblock {\em Theoretical Computer Science}, 943:241--266, 2023.

\bibitem{wang2023regularity}
Shuai Wang, Bingdong Li, and Aimin Zhou.
\newblock A regularity augmented evolutionary algorithm with dual-space search
  for multiobjective optimization.
\newblock {\em Swarm and Evolutionary Computation}, 78:101261, 2023.

\bibitem{hong2020efficient}
Wenjing Hong, Chao Qian, and Ke~Tang.
\newblock Efficient minimum cost seed selection with theoretical guarantees for
  competitive influence maximization.
\newblock {\em IEEE Transactions on Cybernetics}, 51(12):6091--6104, 2020.

\bibitem{liu2021effective}
Shengcai Liu, Ning Lu, Wenjing Hong, Chao Qian, and Ke~Tang.
\newblock Effective and imperceptible adversarial textual attack via
  multi-objectivization.
\newblock {\em arXiv preprint arXiv:2111.01528}, 2021.

\bibitem{liu2022reliable}
Shengcai Liu, Fu~Peng, and Ke~Tang.
\newblock Reliable robustness evaluation via automatically constructed attack
  ensembles.
\newblock In {\em Proceedings of The 37th AAAI Conference on Artificial
  Intelligence}, pages 8852--8860, Washington, DC, 2023.

\bibitem{libingdong2023}
Bingdong Li, Yan Zhang, Peng Yang, Xin Yao, and Aimin Zhou.
\newblock A two-population algorithm for large-scale multi-objective
  optimization based on fitness-aware operator and adaptive environmental
  selection.
\newblock {\em IEEE Transactions on Evolutionary Computation},
  DOI:10.1109/TEVC.2023.3296488, 2023.

\bibitem{chen2021interactive}
Lu~Chen, Bin Xin, and Jie Chen.
\newblock Interactive multi-objective evolutionary algorithm based on
  decomposition and compression.
\newblock {\em Science China Information Sciences}, 64:1--16, 2021.

\bibitem{coello2020ind}
Jes\'{u}s~Guillermo Falc\'{o}n-Cardona and Carlos A.~Coello Coello.
\newblock Indicator-based multi-objective evolutionary algorithms: A
  comprehensive survey.
\newblock {\em ACM Comput. Surv.}, 53(2):35, mar 2020.

\bibitem{deb2002a}
K.~Deb, A.~Pratap, S.~Agarwal, and T.~Meyarivan.
\newblock A fast and elitist multiobjective genetic algorithm: Nsga-ii.
\newblock {\em IEEE Transactions on Evolutionary Computation}, 6(2):182--197,
  2002.

\bibitem{shen2020generating}
Ruimin Shen, Yan Zheng, Jianye Hao, Zhaopeng Meng, Yingfeng Chen, Changjie Fan,
  and Yang Liu.
\newblock Generating behavior-diverse game ais with evolutionary
  multi-objective deep reinforcement learning.
\newblock In {\em IJCAI}, pages 3371--3377, 2020.

\bibitem{song2022evolutionary}
Fuhong Song, Huanlai Xing, Xinhan Wang, Shouxi Luo, Penglin Dai, Zhiwen Xiao,
  and Bowen Zhao.
\newblock Evolutionary multi-objective reinforcement learning based trajectory
  control and task offloading in uav-assisted mobile edge computing.
\newblock {\em IEEE Transactions on Mobile Computing},
  10.1109/TMC.2022.3208457, 2022.

\bibitem{xu2020prediction}
Jie Xu, Yunsheng Tian, Pingchuan Ma, Daniela Rus, Shinjiro Sueda, and Wojciech
  Matusik.
\newblock Prediction-guided multi-objective reinforcement learning for
  continuous robot control.
\newblock In {\em International conference on machine learning}, pages
  10607--10616. PMLR, 2020.

\bibitem{abels2019dynamic}
Axel Abels, Diederik Roijers, Tom Lenaerts, Ann Now{\'e}, and Denis
  Steckelmacher.
\newblock Dynamic weights in multi-objective deep reinforcement learning.
\newblock In {\em International conference on machine learning}, pages 11--20.
  PMLR, 2019.

\bibitem{van2014multi}
Kristof Van~Moffaert and Ann Now{\'e}.
\newblock Multi-objective reinforcement learning using sets of pareto
  dominating policies.
\newblock {\em The Journal of Machine Learning Research}, 15(1):3483--3512,
  2014.

\bibitem{kaushik2018multi}
Rituraj Kaushik, Konstantinos Chatzilygeroudis, and Jean-Baptiste Mouret.
\newblock Multi-objective model-based policy search for data-efficient learning
  with sparse rewards.
\newblock In {\em Proceedings of The 2nd Conference on Robot Learning},
  volume~87, pages 839--855, 2018.

\bibitem{mao2019variance}
Hongzi Mao, Shaileshh~Bojja Venkatakrishnan, Malte Schwarzkopf, and Mohammad
  Alizadeh.
\newblock Variance reduction for reinforcement learning in input-driven
  environments.
\newblock In {\em International Conference on Learning Representations}, 2019.

\bibitem{bian2021robustness}
Chao Bian, Chao Qian, Yang Yu, and Ke~Tang.
\newblock On the robustness of median sampling in noisy evolutionary
  optimization.
\newblock {\em Science China Information Sciences}, 64:1--13, 2021.

\bibitem{hao2023enhancing}
Hao Hao, Xiaoqun Zhang, and Aimin Zhou.
\newblock Enhancing saeas with unevaluated solutions: A case study of relation
  model for expensive optimization.
\newblock {\em arXiv preprint arXiv:2309.11994}, 2023.

\bibitem{liu20213howgood}
Shengcai Liu, Yu~Zhang, Ke~Tang, and Xin Yao.
\newblock How good is neural combinatorial optimization? a systematic
  evaluation on the traveling salesman problem.
\newblock {\em IEEE Computational Intelligence Magazine}, 18(3):14--28, 2023.

\end{thebibliography}



\end{document}